\pdfoutput=1
\documentclass{article}
\usepackage{arxiv}
\usepackage[T1]{fontenc}
\usepackage{amsmath,amsfonts}
\usepackage{algorithmic}
\usepackage{algorithm}
\usepackage{array}
\usepackage[font=small,labelfont=bf]{subfig}
\usepackage{textcomp}

\usepackage{url}
\usepackage{verbatim}
\usepackage{graphicx}
\usepackage{cite}
\usepackage{booktabs}

\usepackage{multirow}
\usepackage{amssymb}

\usepackage{bm}

\usepackage[hidelinks]{hyperref}
\hypersetup{pdftitle={Spectral Consistency-Guided Multiview Point Cloud Registration for Low-Overlap Scenes},pdfauthor={Tianyu Li, Yanghong Lin, Shudong Zhou, Kui Yang, Jingru Zhang, Li Fang, Wei Yao}}
\renewcommand{\shorttitle}{Spectral Consistency-Guided Multiview Registration}
\date{}
\begin{document}

\title{Spectral Consistency-Guided Multiview Point Cloud Registration for Low-Overlap Scenes}

\author{%
\begin{tabular}{c}
Tianyu Li$^{1,2}$, Yanghong Lin$^{2,3}$, Shudong Zhou$^{1,2}$, Kui Yang$^1$,\\
Jingru Zhang$^{2,3}$, Li Fang$^{2,3}$, Wei Yao$^{2,3}$
\end{tabular}%
\thanks{This work was supported by the Jing-Jin-Ji Regional Integrated Environmental Improvement-National Science and Technology Major Project (2026ZD1214400). Corresponding authors: Li Fang and Wei Yao. E-mail: \href{mailto:wyao@iue.ac.cn}{wyao@iue.ac.cn}.}\\[0.7em]
\small $^1$School of Resources and Environmental Sciences, Wuhan University, Wuhan, China\\
\small $^2$Spatial Intelligence and Urban Computing, Institute of Urban Environment,\\
\small Chinese Academy of Sciences, Xiamen, China\\
\small $^3$State Key Lab for Ecological Security of Regions and Cities,\\
\small Institute of Urban Environment, Chinese Academy of Sciences, Xiamen, China
}

\maketitle

\begin{abstract}
Multiview point cloud registration is particularly challenging in low-overlap scenes, where reliable correspondences are limited and incorrect pairwise transformations can affect global pose estimation. In addition, registering all scan pairs is computationally expensive because many pairs provide weak geometric information. To address these problems, we propose GMPCR, a non-learning-based spectral consistency-guided framework for efficient and robust multiview point cloud registration. GMPCR builds a refined second-order compatibility structure from initial correspondences and uses its dominant spectral response to evaluate both correspondence reliability and scan-pair confidence. This allows unreliable correspondences to be filtered and informative scan pairs to be selected before relative transformation estimation, leading to a sparse pose graph and reduced pairwise registration cost. For each retained scan pair, maximal-clique-based hypothesis generation is used to estimate reliable relative transformations. The resulting pose graph is further refined by an adaptive history-aware synchronization scheme, in which the effect of residual history is adjusted according to changes in the global rotation residual. A recovery mechanism also allows down-weighted edges to regain confidence when their global consistency improves. Experiments on 3DMatch, 3DLoMatch, ScanNet, and ETH demonstrate the effectiveness of GMPCR. It achieves registration recalls of 97.2\% and 89.6\% on 3DMatch and 3DLoMatch, respectively, while maintaining competitive performance on ScanNet and ETH. The results show that GMPCR provides a favorable balance among registration accuracy, robustness to low overlap, and computational efficiency. The code is publicly available at \url{https://github.com/swccj/gmpcr}.

\end{abstract}


\section{Introduction}

Point cloud registration is a fundamental task in 3D computer vision, with broad applications in robotic navigation~\cite{li2026mapping,hu2025semantic}, autonomous driving\cite{ye2025fade3d,zhao2025safety}, 3D reconstruction\cite{mahmoud2026indoor,fang2025coupled}, and augmented reality\cite{peek2025novel}. Given a set of partially overlapping scans, multiview point cloud registration aims to estimate a globally consistent pose for each scan to align all observations within a common coordinate frame. Compared with pairwise registration \cite{zhang2026deep,huang2021robust,xu2019pairwise,polewski2019scale}, the multiview problem introduces an additional challenge: reliable overlap connections between scans must be identified while their relative transformations are jointly reconciled into a globally consistent configuration. This problem becomes particularly challenging under low overlap. Restricted viewpoints, occlusions, and limited sensor coverage may leave only a small common region between two scans, resulting in very few correct correspondences among a large number of ambiguous or incorrect matches \cite{huang2021predator, chen2022sc2pcr,huang2021pairwise}. At the same time, the overlap between scans is generally unknown in advance. For $N$ scans, the number of possible scan pairs grows quadratically, although only a small fraction may provide useful geometric constraints. Exhaustively performing full registration on all candidate pairs is therefore computationally inefficient and may introduce unreliable transformations from weakly overlapping or non-overlapping scan pairs into the pose graph \cite{wang2023robust,vedrenne2025polar}.

Existing approaches address different aspects of this problem. Robust pairwise registration methods improve transformation estimation through learned descriptors\cite{FCGF2019}, geometric compatibility\cite{1544893}, correspondence pruning\cite{bai2021pointdsc}, or consensus-based hypothesis generation\cite{zhang2023mac}, while multiview methods improve global consistency through pose graph construction\cite{zhao2023registration}, robust synchronization\cite{wang2023robust}, and iterative edge reweighting\cite{fang2024robust}. However, correspondence reliability, scan-pair selection, and global pose refinement are often handled separately. As a result, computationally expensive relative transformation estimation may be performed before the reliability of a scan pair is established. This issue becomes more serious in low-overlap scenes, where correct correspondences are scarce and only a limited number of reliable scan connections may be available.

An effective multiview registration framework should therefore consider reliability at multiple levels. First, informative scan pairs should be identified before expensive relative transformation estimation. Second, correct correspondences within each retained pair should be distinguished from outliers using reliable geometric consistency cues. Third, relative transformations that remain inconsistent with the global pose configuration should be gradually down-weighted. Achieving these objectives simultaneously is challenging. Aggressive filtering may discard the limited geometric information available under low overlap, whereas retaining too many scan pairs, correspondences, or transformation hypotheses increases both computational cost and the risk of introducing outliers. In particular, correspondence-level geometric information is typically used only to evaluate matches within individual scan pairs, while pose graph construction relies on separate overlap or connectivity measures to select scan relations. Consequently, the geometric structure of the correspondences is not fully used to determine, before relative pose estimation, whether a scan pair should be included in the pose graph. Moreover, residual-based pose graph reweighting must balance rapid adaptation to changes in the global estimates with stable refinement near convergence, especially when reliable edges temporarily exhibit large residuals during the early iterations.

To address these challenges, we propose GMPCR, a spectral consistency-guided framework for reliable and efficient multiview point cloud registration. GMPCR first constructs first- and second-order compatibility structures from the initial correspondences of each candidate scan pair. The dominant spectral response of the refined compatibility structure is used not only to evaluate correspondence reliability but also to estimate scan-pair confidence. This allows informative scan pairs to be selected before relative transformation estimation, so that pairwise registration is performed only on reliable scan connections. For each retained scan pair, maximal-clique-based hypothesis generation is used to estimate the relative transformation, resulting in a sparse weighted pose graph.

Local geometric reliability alone cannot ensure consistency across the entire pose graph. We therefore introduce an adaptive history-aware synchronization strategy that updates edge weights according to the current residuals and the evolution of the global solution. The influence of residual history is reduced when the global rotation residual changes significantly and increased as the optimization becomes stable. In addition, a one-sided recovery mechanism allows previously down-weighted edges to regain confidence when their residuals decrease. In this way, GMPCR connects correspondence-level geometric consistency, scan-pair reliability, and global pose consistency within a unified local-to-global registration framework.

The main contributions of this work are summarized as follows:

\begin{itemize}
    \item We introduce a cross-level spectral reliability measure that uses the dominant structure of a refined second-order correspondence compatibility graph for both correspondence filtering and scan-pair confidence estimation, linking local geometric consistency to pose graph construction.

    \item We develop a scan-pair selection strategy that exploits correspondence-level spectral consistency to identify informative scan connections before relative transformation estimation, avoiding unnecessary pairwise registration and constructing a sparse yet reliable pose graph for global synchronization.

    \item We develop an adaptive history-aware synchronization scheme that adjusts the influence of residual history according to the evolution of the global rotation residual. A one-sided recovery mechanism further allows previously down-weighted edges to regain confidence when their consistency with the global pose configuration improves.
\end{itemize}

Extensive experiments on 3DMatch, 3DLoMatch, ScanNet, and ETH demonstrate that GMPCR achieves a favorable balance between registration accuracy and computational efficiency.

\section{Related Work}

\subsection{Correspondence-based Multiview Point Cloud Registration}
Correspondence-based methods are widely used for multiview point cloud registration. A typical pipeline first establishes correspondences and estimates pairwise transformations, and then recovers globally consistent scan poses through graph optimization, motion averaging, or transformation synchronization~\cite{torsello2011multiview,theiler2015globally}. The explicit geometric relations provided by correspondences and pose graph edges make this framework naturally suited to robust estimation and sparse graph construction. Moreover, correspondence-level geometric consistency can be directly evaluated, providing useful information for both pairwise transformation estimation and the subsequent construction of the pose graph.

Recent methods improve either pairwise reliability or global pose graph consistency. Learning-based approaches exploit correspondence confidence or pairwise relations to improve registration, while LMVR \cite{gojcic2020learning} jointly considers pairwise alignment and multiview refinement. SGHR~\cite{wang2023robust} predicts scan overlap to construct a sparse pose graph and applies history reweighting during synchronization. Bundle-adjustment methods further refine multiple scan poses jointly~\cite{huang2021bundle}. More recently, SMVR~\cite{fang2024robust} evaluates scan relationships using algebraic connectivity and spatial compatibility to select reliable graph connections before transformation estimation. These methods reduce the influence of unreliable pairwise relations and improve global consistency, but their performance still depends strongly on the reliability of the initial scan connections and correspondence sets. However, low-overlap scenes remain challenging because scan-pair reliability must be estimated from correspondence sets containing few inliers and many outliers. Feature similarity alone may provide insufficient geometric cues for distinguishing reliable scan pairs, especially when repetitive structures generate ambiguous matches.

Existing spectral methods mainly focus on correspondence pruning or graph analysis separately. In contrast, GMPCR links correspondence filtering, scan-pair selection, and global synchronization within a unified reliability framework. The refined spectral structure of correspondence-level geometric consistency is used to evaluate both correspondence reliability and scan-pair confidence before relative transformation estimation. This enables the same geometric information to guide correspondence filtering and pose graph construction, while the resulting pairwise transformations are further refined through adaptive history-aware synchronization based on their consistency with the global pose configuration.

\subsection{Correspondence-free Multiview Point Cloud Registration}

Correspondence-free registration avoids explicit point-to-point matching by aligning multiple scans through distribution-level, continuous-function, latent-space, or map-level objectives. Probabilistic methods jointly estimate scan poses and shared mixture models~\cite{evangelidis2014generative,evangelidis2018joint,zhang2021lmm,pan2025ejrgf}, while DeepMapping and DeepMapping2 formulate multiview registration as map optimization~\cite{ding2019deepmapping,chen2023deepmapping2}. RKHS-based methods align continuous geometric or semantic representations~\cite{clark2021nonparametric,zhang2021semanticcvo,zhang2025rkhsba}, and other approaches optimize depth-based scene representations~\cite{zhou2025depthguided} or latent features~\cite{vedrenne2025polar}. These formulations reduce the dependence on discrete correspondence construction and provide flexible ways to exploit dense geometric or learned scene representations. More recently, RAP~\cite{pan2026eccv} uses flow matching to directly generate jointly registered point clouds and subsequently recover rigid poses, providing an efficient correspondence-free alternative without explicit pairwise pose graph construction. This generative formulation further demonstrates the potential of directly modeling joint multiview alignment instead of decomposing the problem into independent pairwise registration tasks.

Despite their advantages, correspondence-free methods generally rely on global scene or representation-level objectives and do not explicitly model scan-pair reliability or correspondence-level geometric information. Their performance may therefore be affected when global representations are unreliable, initialization errors are large, or only limited overlap is available between scans. Moreover, the absence of explicit pairwise geometric relations makes it less straightforward to identify informative scan connections before global optimization. This makes such methods less suitable for explicit sparse pose graph construction and direct control of pairwise computation. In contrast, our method retains a correspondence-based formulation and exploits spectral geometric consistency to select informative scan pairs before transformation estimation, filter unreliable correspondences, and construct a sparse pose graph for adaptive global synchronization.

\section{Method}

\subsection{Overview}

Given $N$ point cloud scans $\mathcal{P}=\{\mathbf{P}_i\}_{i=1}^{N}$, multiview registration aims to estimate a global rigid transformation $\mathbf{T}_i\in\mathrm{SE}(3)$ for each scan, mapping all scans into a common coordinate frame. In practice, unknown scan overlap and highly contaminated initial correspondences make exhaustive pairwise registration computationally expensive and prone to unreliable pose graph edges.

\begin{figure}[htbp]
    \centering
    \includegraphics[width=\linewidth]{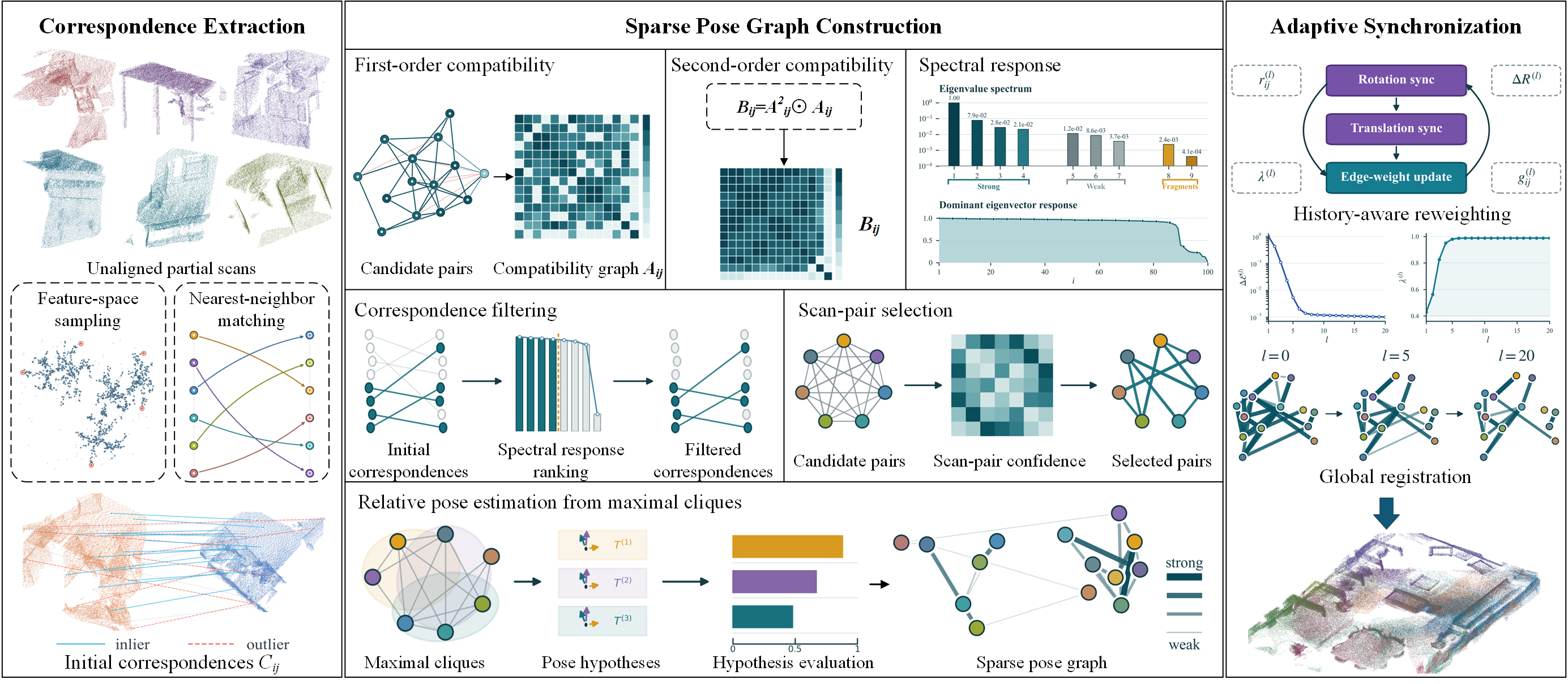}
    \caption{Overview of the proposed GMPCR framework.}
    \label{fig:pipeline}
\end{figure}

The overall pipeline of GMPCR is illustrated in Fig.~\ref{fig:pipeline} and consists of two stages. The first stage performs \emph{spectral consistency-guided pairwise registration}. Initial correspondences are established by feature sampling and mutual nearest-neighbor matching. Compatibility analysis is then used to evaluate correspondence reliability and scan-pair confidence, retaining only informative scan pairs. For each selected pair, geometrically consistent maximal cliques generate transformation hypotheses through weighted rigid alignment, and the hypothesis with the largest inlier support is selected. This stage yields a sparse weighted pose graph $\mathcal{G}_p=(\mathcal{V},\mathcal{E}_p)$ with estimated relative transformations $\widehat{\mathbf{T}}_{ij}$ and initial edge weights $w_{ij}^{(0)}$.

The second stage performs \emph{adaptive history-aware pose synchronization}. Global poses and edge weights are iteratively refined according to global consistency. Residual history stabilizes the weight updates, while a recovery mechanism allows previously down-weighted edges to regain confidence when their consistency improves. The final refined pose graph is used to recover globally consistent poses $\{\widehat{\mathbf{T}}_i\}_{i=1}^{N}$.

\subsection{Spectral Consistency-Guided Pairwise Registration}

The pairwise registration stage constructs a sparse set of reliable relative transformations for subsequent global synchronization. Instead of estimating a transformation for every possible scan pair, GMPCR progressively converts feature similarity into correspondence-level geometric consistency and then into pose graph edges. This stage consists of three steps: initial correspondence construction, spectral consistency-guided scan pair selection, and clique-based transformation estimation.

\subsubsection{Initial Correspondence Construction}

For scan $\mathbf{P}_i$, let $\{(\mathbf{x}_a^i,\mathbf{f}_a^i)\}_{a=1}^{N_i}$ denote its points and associated local descriptors. Directly matching all points introduces considerable redundancy and increases the cost of processing all unordered scan pairs. GMPCR therefore performs farthest-point sampling in the feature space and retains $N_s$ representative points. Compared with spatial sampling, feature-space sampling promotes descriptor diversity and reduces the dominance of densely distributed and highly similar descriptors.

For each scan pair $(i,j)$, nearest-neighbor searches are conducted between the sampled descriptor sets. A match is retained only when its two descriptors are mutual nearest neighbors. The retained matches are ranked in ascending order of feature distance, and $N_o$ matches are kept. The resulting initial correspondence set is
\begin{equation}
    \mathcal{C}_{ij}
    =
    \left\{
        c_m^{ij}:=(\mathbf{x}_m^i,\mathbf{x}_m^j)
    \right\}_{m=1}^{N_o}.
\end{equation}
Mutual matching suppresses asymmetric and many-to-one assignments, while limiting the number of feature-nearest matches bounds the size of the subsequent compatibility graph. Nevertheless, feature similarity alone is insufficient to guarantee geometric correctness, particularly in repetitive structures and low-overlap regions. We therefore treat $\mathcal{C}_{ij}$ as a candidate correspondence set whose reliability is subsequently assessed through collective geometric consistency.

\subsubsection{Spectral Consistency-Guided Scan Pair Selection}

For each correspondence set $\mathcal{C}_{ij}$, GMPCR first constructs a first-order compatibility graph (FCG)\cite{1544893}, in which each node represents a correspondence. Given two correspondences $c_m^{ij}$ and $c_n^{ij}$, their rigidity discrepancy is defined as
\begin{equation}
    \delta_{mn}^{ij}
    =
    \left|
        \|\mathbf{x}_m^i-\mathbf{x}_n^i\|_2
        -
        \|\mathbf{x}_m^j-\mathbf{x}_n^j\|_2
    \right|.
\end{equation}
Because a rigid transformation preserves pairwise Euclidean distances, a small $\delta_{mn}^{ij}$ indicates that the two correspondences are geometrically compatible. Their preliminary affinity is defined as
\begin{equation}
    \bar{a}_{mn}^{ij}
    =
    \max\left(
        0,
        1-\frac{(\delta_{mn}^{ij})^2}{\tau_c^2}
    \right).
\end{equation}
The sparsified FCG affinity is
\begin{equation}
    a_{mn}^{ij}
    =
    \begin{cases}
        \bar{a}_{mn}^{ij},
        & m\neq n\\
        0, & \text{otherwise},
    \end{cases}
\end{equation}
where $\tau_c>0$ controls the geometric tolerance. The corresponding weighted adjacency matrix is
\begin{equation}
    \mathbf{A}_{ij}
    :=
    [a_{mn}^{ij}]_{m,n=1}^{N_o}
\end{equation}
This sparsification reduces the influence of weak or noisy geometric relations on the subsequent spectral analysis.

Pairwise compatibility alone may still admit accidental agreements, especially in scenes containing repeated structures. Following the second-order spatial compatibility formulation in~\cite{chen2022sc2pcr}, we construct the second-order compatibility graph (SCG):
\begin{equation}
    \mathbf{B}_{ij}
    =
    \mathbf{A}_{ij}^{2}\odot\mathbf{A}_{ij},
\end{equation}
where $\odot$ denotes the Hadamard product. Its entry
\begin{equation}
    b_{mn}^{ij}
    :=
    [\mathbf{B}_{ij}]_{mn}
    =
    a_{mn}^{ij}
    \sum_{l=1}^{N_o}a_{ml}^{ij}a_{ln}^{ij}
\end{equation}
combines the direct compatibility between $c_m^{ij}$ and $c_n^{ij}$ with their shared-neighbor support. A large $b_{mn}^{ij}$ therefore indicates strong first- and second-order geometric consistency. Compared with pairwise distance compatibility alone, the second-order structure is more robust to false geometric matches because an incorrect correspondence is less likely to receive consistent support from multiple neighboring correspondences. Building on this structure, GMPCR further exploits its refined spectral information for both correspondence pruning and scan-pair confidence estimation before relative transformation estimation.

\begin{figure}[htbp]
    \centering
    \includegraphics[width=\textwidth]{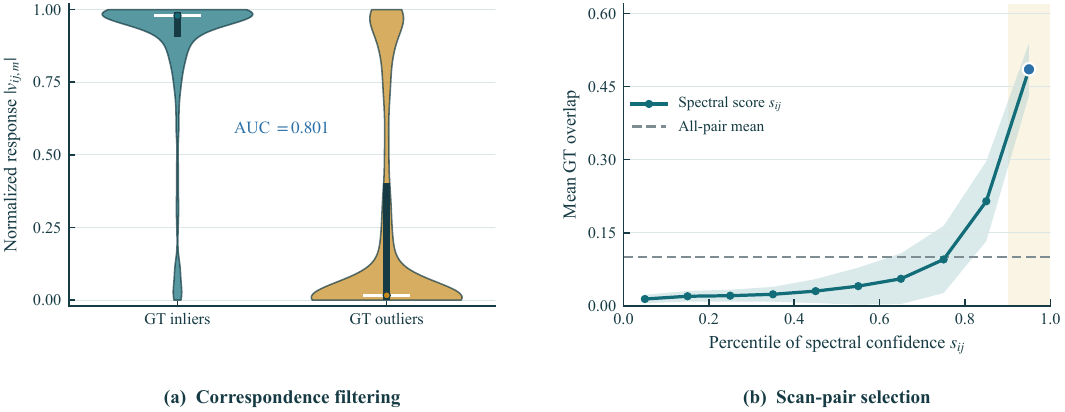}
    \caption{Spectral evidence for the two selection stages of GMPCR. (a) Correspondence filtering ranks putative correspondences according to the magnitudes of the dominant-eigenvector components of the compatibility graph. Ground-truth inliers are concentrated near a response of one, whereas outliers are mainly distributed near zero, yielding a mean pairwise AUC of 0.801 over 3DLoMatch scan pairs. (b) Scan pairs are grouped into deciles by spectral confidence and ordered from low to high along the horizontal axis, while the vertical axis reports the mean ground-truth overlap in each decile. The teal curve shows an overall increase in overlap with confidence. The gray dashed line marks the mean overlap over all candidate pairs (approximately 0.10), and the light-cyan band indicates the 95\% confidence interval of the scene-averaged overlap across the eight test scenes. The yellow region highlights the highest-confidence decile, whose mean overlap is approximately 0.485, or $4.8\times$ the all-pair mean.}
    \label{fig:spectral-selection-evidence}
\end{figure}

Let $\mathbf{v}_{ij}\in\mathbb{R}^{N_o}$ be the unit-norm dominant eigenvector of $\mathbf{B}_{ij}$, with $v_{ij,m}$ denoting its $m$-th entry. The magnitude $|v_{ij,m}|$ reflects the participation of correspondence $c_m^{ij}$ in the dominant geometrically coherent structure. As illustrated in Fig.~\ref{fig:spectral-selection-evidence}(a), ground-truth inliers
generally exhibit stronger spectral responses than outliers. GMPCR retains the top $r_s$ fraction of correspondences according to $|v_{ij,m}|$. Let $\mathcal{S}_{ij}$ be the retained index set and define
\begin{equation}
    \widetilde{\mathcal{C}}_{ij}
    :=
    \{c_m^{ij}\mid m\in\mathcal{S}_{ij}\},
    \qquad
    \widetilde{M}_{ij}
    :=
    |\widetilde{\mathcal{C}}_{ij}|.
\end{equation}
For notational simplicity, the retained correspondences are reindexed from $1$ to $\widetilde{M}_{ij}$, while their point coordinates continue to be denoted by $(\mathbf{x}_m^i,\mathbf{x}_m^j)$. The FCG is then restricted to the retained indices, after which the SCG and its dominant eigenvector are recomputed:
\begin{equation}
    \widetilde{\mathbf{A}}_{ij}
    =
    [\mathbf{A}_{ij}]_{\mathcal{S}_{ij},\mathcal{S}_{ij}},
    \qquad
    \widetilde{\mathbf{B}}_{ij}
    =
    \widetilde{\mathbf{A}}_{ij}^{2}
    \odot
    \widetilde{\mathbf{A}}_{ij},
\end{equation}
where $\widetilde{\mathbf{v}}_{ij}\in\mathbb{R}^{\widetilde{M}_{ij}}$ denotes the unit-norm dominant eigenvector of $\widetilde{\mathbf{B}}_{ij}$. This two-pass construction removes weak correspondences before rebuilding the neighborhood structure, providing a refined graph for both scan-pair scoring and transformation estimation. Recomputing the graph ensures that discarded correspondences no longer contribute to common-neighbor support once they are removed from registration.

To estimate the reliability of each candidate scan pair, we define a scan-pair confidence directly from the refined correspondence compatibility structure. Unlike our previous SMVR framework~\cite{fang2024robust}, which evaluates scan relationships using algebraic connectivity and spatial compatibility at the pose graph level, GMPCR computes the dominant spectral response from the refined second-order correspondence graph. In this way, scan-pair confidence is determined directly by the geometric consistency of the underlying correspondences. Moreover, the confidence is evaluated before relative transformation estimation and is used to select the scan pairs for subsequent registration. Specifically, the raw geometric confidence of scan pair $(i,j)$ is defined by the Rayleigh quotient:
\begin{equation}
\widehat{s}_{ij}
=
\widetilde{\mathbf{v}}_{ij}^{\top}
\widetilde{\mathbf{B}}_{ij}
\widetilde{\mathbf{v}}_{ij}.
\end{equation}
The raw confidence scores of all candidate scan pairs are normalized to obtain the final scan-pair confidence:
\begin{equation}
s_{ij}
=
\frac{\widehat{s}_{ij}}
{\|\widehat{\mathbf{s}}\|_2},
\end{equation}
where $\widehat{\mathbf{s}}$ denotes the vector containing the raw confidence scores of all unordered candidate scan pairs. When $\|\widehat{\mathbf{s}}\|_2=0$, all confidence scores are set to zero.

Since $\widetilde{\mathbf{B}}_{ij}$ is a nonnegative matrix, the dominant eigenvector can be selected to be nonnegative according to the Perron--Frobenius theorem, ensuring $\widehat{s}_{ij}\geq0$. Moreover, when the dominant eigenvector is computed exactly, the Rayleigh quotient equals the largest eigenvalue of $\widetilde{\mathbf{B}}_{ij}$.

The proposed spectral confidence measures the intrinsic geometric coherence of each scan pair. As shown in Fig.~\ref{fig:spectral-selection-evidence}(b), scan pairs with higher spectral confidence generally exhibit larger ground-truth overlap. Reliable scan pairs usually contain correspondences that form compact and mutually consistent structures within the refined second-order compatibility graph, producing stronger spectral responses. In contrast, unreliable scan pairs tend to exhibit fragmented compatibility patterns with isolated or weakly supported correspondences, resulting in lower confidence values. Therefore, the proposed confidence captures both direct correspondence compatibility and higher-order consistency encoded by the refined second-order graph, providing a reliable criterion for selecting edges during pose graph construction.

For scan $i$, let $\mathcal{N}_K(i)$ denote the set of its $K$ highest-confidence neighboring scans. The candidate pose graph edge set is formed as the union of the directed top-$K$ neighborhoods:
\begin{equation}
    \mathcal{E}_p
    =
    \left\{
        (i,j)
        \;\middle|\;
        i<j,\ 
        j\in\mathcal{N}_K(i)
        \ \text{or}\ 
        i\in\mathcal{N}_K(j)
    \right\}.
\end{equation}
This union rule retains an edge if either scan identifies the other as a reliable neighbor, preserving informative connections while maintaining a sparse pose graph.

\subsubsection{Clique-Based Transformation Estimation}

For each retained scan pair $(i,j)\in\mathcal{E}_p$, we adopt maximal-clique-based hypothesis generation~\cite{zhang2023mac} for pairwise transformation estimation. This stage is performed after spectral scan-pair selection, so relative pose estimation is applied only to the retained scan connections. An unweighted graph is constructed from the refined compatibility structure, and maximal cliques with at least three nodes are enumerated. Each clique defines a geometrically consistent correspondence subset used to generate a transformation hypothesis.

Since a compatibility graph may contain many maximal cliques, GMPCR ranks them by their mean internal SCG affinity:
\begin{equation}
    q_{ij}(\mathcal{M})
    =
    \frac{1}{|\mathcal{M}|(|\mathcal{M}|-1)}
    \sum_{\substack{m,n\in\mathcal{M}\\m\neq n}}
    \widetilde{b}_{mn}^{ij}.
\end{equation}
A larger $q_{ij}(\mathcal{M})$ indicates stronger geometric consistency within the clique. Only the top $K_c$ cliques are retained, and their collection is denoted by $\mathcal{H}_{ij}$.
Using the mean affinity rather than the total affinity reduces the preference for large cliques and allows smaller but highly consistent subsets to remain competitive in low-overlap cases.

Each clique $\mathcal{M}\in\mathcal{H}_{ij}$ generates one relative transformation hypothesis. The associated rotation and translation are estimated by
\begin{equation}
\begin{aligned}
    (\widehat{\mathbf{R}}_{ij}^{\mathcal{M}},
     \widehat{\mathbf{t}}_{ij}^{\mathcal{M}})
    =
              {\arg\min}
    \sum_{m\in\mathcal{M}}
    |\widetilde{v}_{ij,m}|
    \left\|
        \mathbf{x}_m^i
        -(\mathbf{R}\mathbf{x}_m^j+\mathbf{t})
    \right\|_2^2.
\end{aligned}
\end{equation}
Each hypothesis is evaluated on the complete filtered correspondence set. Given a candidate transformation $\mathbf{T}=(\mathbf{R},\mathbf{t})$, its pair-specific inlier ratio is defined as
\begin{equation}
    \operatorname{IR}_{ij}(\mathbf{T})
    =
    \frac{1}{\widetilde{M}_{ij}}
    \sum_{m=1}^{\widetilde{M}_{ij}}
    \mathbb{I}\!\left(
        \left\|
            \mathbf{x}_m^i
            -(\mathbf{R}\mathbf{x}_m^j+\mathbf{t})
        \right\|_2
        <\tau_{\mathrm{in}}
    \right),
\end{equation}
where $\tau_{\mathrm{in}}$ is the inlier distance threshold and $\mathbb{I}(\cdot)$ denotes the indicator function. The relative transformation $\widehat{\mathbf{T}}_{ij}$ with the highest inlier ratio is selected as the final transformation. The initial weight of pose graph edge $(i,j)$ is defined as
\begin{equation}
    w_{ij}^{(0)}
    =
    s_{ij}\operatorname{IR}_{ij}(\widehat{\mathbf{T}}_{ij}).
\end{equation}
The inlier ratio measures the support of the selected transformation over the filtered correspondence set. The pairwise registration stage therefore outputs
\begin{equation}
    \left\{
        (\widehat{\mathbf{T}}_{ij},w_{ij}^{(0)})
    \right\}_{(i,j)\in\mathcal{E}_p},
\end{equation}
which defines the sparse weighted pose graph used for subsequent global synchronization.

\subsection{Adaptive History-aware Pose Synchronization}

The pairwise registration stage produces a sparse weighted pose graph $\mathcal{G}_p=(\mathcal{V},\mathcal{E}_p)$, where each edge $(i,j)\in\mathcal{E}_p$ contains an estimated relative transformation $\widehat{\mathbf{T}}_{ij}$ and an initial weight $w_{ij}^{(0)}$. Pose synchronization seeks global transformations that agree with these pairwise measurements. Although spectral correspondence filtering removes many unreliable scan pairs and correspondences, a small number of incorrect transformations may remain in the pose graph. Under low-overlap conditions, some outlier edges can receive relatively large initial weights and bias the synchronized poses.

To reduce the influence of unreliable edges, GMPCR introduces an adaptive history-aware iteratively reweighted least squares (IRLS) scheme. The contribution of residual history is adjusted according to the evolution of the global solution, and a recovery mechanism is introduced to restore confidence for edges whose global consistency improves. Each IRLS iteration alternates between pose synchronization and edge-weight updates. The synchronization step estimates the global rotations and translations using the current edge weights, while the reweighting step evaluates the consistency of each relative rotation with the current global poses and updates its confidence. Let $\ell\in{0,\ldots,L-1}$ denote the iteration index, where $L$ is the maximum number of IRLS iterations, and let $w_{ij}^{(\ell)}$ denote the weight of edge $(i,j)$ at iteration $\ell$. The weights are initialized using the pairwise registration confidence ${w_{ij}^{(0)}}$.

\subsubsection{Rotation and Translation Synchronization}

Given the current edge weights $\{w_{ij}^{(\ell)}\}_{(i,j)\in\mathcal{E}_p}$ and relative transformations $\{\widehat{\mathbf{T}}_{ij}\}_{(i,j)\in\mathcal{E}_p}$, pose synchronization recovers a globally consistent transformation for each scan. Following the standard synchronization procedure adopted in SGHR~\cite{wang2023robust}, we first estimate the global rotations and subsequently solve for the global translations under the recovered rotations.

At IRLS iteration $\ell$, the global rotations are estimated by
\begin{equation}
    \{\mathbf{R}_i^{(\ell)}\}_{i=1}^{N}
    =
    \underset{\{\mathbf{R}_i\in\mathrm{SO}(3)\}_{i=1}^{N}}
              {\arg\min}
    \sum_{(i,j)\in\mathcal{E}_p}
    w_{ij}^{(\ell)}
    \left\|
        \widehat{\mathbf{R}}_{ij}
        -\mathbf{R}_i^{\top}\mathbf{R}_j
    \right\|_{\mathrm{F}}^{2},
\end{equation}
where $\widehat{\mathbf{R}}_{ij}$ is the rotation component of $\widehat{\mathbf{T}}_{ij}$ and $\|\cdot\|_{\mathrm{F}}$ denotes the Frobenius norm. The objective enforces agreement between each measured relative rotation and the relative rotation induced by the global estimates. It is solved by constructing a weighted connection Laplacian, extracting the eigenvectors associated with its three smallest eigenvalues, and projecting the corresponding $3\times3$ blocks onto $\mathrm{SO}(3)$. Consequently, high-confidence edges contribute more strongly to the synchronized rotations, whereas unreliable edges exert less influence.

Once the rotations are determined, the global translations are estimated by
\begin{equation}
    \{\mathbf{t}_i^{(\ell)}\}_{i=1}^{N}
    =
    \underset{\{\mathbf{t}_i\in\mathbb{R}^{3}\}_{i=1}^{N}}
              {\arg\min}
    \sum_{(i,j)\in\mathcal{E}_p}
    w_{ij}^{(\ell)}
    \left\|
        \mathbf{R}_i^{(\ell)}\widehat{\mathbf{t}}_{ij}
        +\mathbf{t}_i-\mathbf{t}_j
    \right\|_2^2,
\end{equation}
where $\widehat{\mathbf{t}}_{ij}$ is the translation component of $\widehat{\mathbf{T}}_{ij}$. This objective follows from $\widehat{\mathbf{T}}_{ij}\approx\mathbf{T}_i^{-1}\mathbf{T}_j$ and is solved by weighted linear least squares.

\subsubsection{Adaptive History-aware Edge-Weight Update}

Pose synchronization and edge reweighting play complementary roles within IRLS. The synchronization step estimates the global poses under the current edge weights, while the reweighting step evaluates the agreement between each measured relative rotation and the current global configuration. During the early iterations, the global pose estimates may still be inaccurate. As a result, a reliable edge can temporarily produce a large residual, while an unreliable edge may appear consistent with the current solution. Updating edge weights based only on the current residual can therefore lead to incorrect weight changes. To reduce this effect, GMPCR maintains a residual history for each edge and adjusts its influence according to the evolution of the global rotation residual. This provides faster response when the global solution changes and more stable weight updates as the optimization approaches convergence.

The normalized rotation residual of edge $(i,j)$ is
\begin{equation}
r_{ij}^{(\ell)}
=
\frac{1}{s_r}
d_{\mathrm{R}}
\left(
\mathbf{R}_{ij}^{(\ell)},
\widehat{\mathbf{R}}_{ij}
\right),
\end{equation}
where $d_{\mathrm{R}}(\cdot,\cdot)$ denotes the angular distance between two rotations and $s_r>0$ is the residual normalization factor. A small residual indicates agreement between the measured relative rotation and the current global configuration, whereas a large residual suggests that the edge is unreliable or that the current solution remains unstable. The overall rotation error at iteration $\ell$ is measured by
\begin{equation}
\mathcal{R}^{(\ell)}
=
\sum_{(i,j)\in\mathcal{E}_p}
w_{ij}^{(\ell)}
\left(r_{ij}^{(\ell)}\right)^2.
\end{equation}
This quantity measures the aggregate weighted discrepancy between the pairwise rotations and the current global solution. Because edges that have already been down-weighted contribute less to the error measure, higher-confidence edges continue to dominate the evaluation of the current optimization state. The relative variation of the overall rotation error is defined as
\begin{equation}
\Delta \mathcal{R}^{(\ell)}
=
\frac{
\left|\mathcal{R}^{(\ell)}-\mathcal{R}^{(\ell-1)}\right|
}{
\left|\mathcal{R}^{(\ell-1)}\right|
}.
\end{equation}
Compared with the absolute error magnitude, $\Delta\mathcal{R}^{(\ell)}$ more directly characterizes the evolution of the synchronization process and is less sensitive to differences in graph size and initial weight scale.

A fixed memory coefficient cannot balance rapid adaptation in early iterations and stable refinement near convergence. Excessive historical influence may preserve inaccurate early residuals, whereas insufficient smoothing near convergence can make the edge weights overly sensitive to small residual variations. GMPCR therefore maintains an exponentially weighted residual history $h_{ij}^{(\ell)}$ with an adaptive memory coefficient $\lambda^{(\ell)}$:
\begin{equation}
h_{ij}^{(\ell)}
=
\lambda^{(\ell)}h_{ij}^{(\ell-1)}
+
\left(1-\lambda^{(\ell)}\right)r_{ij}^{(\ell)}.
\end{equation}
The adaptive memory coefficient is defined as
\begin{equation}
\lambda^{(\ell)}
=
\lambda_{\min}
+(\lambda_{\max}-\lambda_{\min})
\exp\!\left(-\kappa\Delta\mathcal{R}^{(\ell)}\right),
\end{equation}
where $0\leq\lambda_{\min}\leq\lambda_{\max}<1$ define the lower and upper bounds of the memory coefficient and $\kappa>0$ controls its sensitivity to the relative variation of the overall rotation error. When the overall rotation error changes substantially, a large $\Delta\mathcal{R}^{(\ell)}$ moves $\lambda^{(\ell)}$ toward $\lambda_{\min}$, increasing the contribution of the current residual and allowing the history to respond rapidly. As the solution stabilizes, $\lambda^{(\ell)}$ approaches $\lambda_{\max}$, providing stronger historical smoothing and reducing small weight variations. However, residual history may respond slowly when a previously down-weighted edge becomes consistent with the evolving global solution. GMPCR introduces a recovery term
\begin{equation}
g_{ij}^{(\ell)}
=
\max\left(
0,
r_{ij}^{(\ell-1)}-r_{ij}^{(\ell)}
\right).
\end{equation}
The recovery term is positive only when the edge residual decreases, indicating improved consistency with the current global solution. Edges with non-decreasing residuals receive no recovery, while persistent outliers are unlikely to recover substantially because their residual histories remain large. Since the updated weight is capped by the initial pairwise confidence, the recovery mechanism cannot increase an edge weight beyond its initial value. It therefore restores confidence only when supported by improved global consistency. Combining the residual history and recovery term yields the updated edge weight
\begin{equation}
w_{ij}^{(\ell+1)}
=
w_{ij}^{(0)}
\exp\!\left(
-h_{ij}^{(\ell)}
+\rho g_{ij}^{(\ell)}
\right),
\end{equation}
where $\rho\geq0$ controls the recovery strength. To prevent excessive weight growth, the updated edge weight is capped by its initial confidence. Thus, the recovery mechanism only restores previously reduced confidence without exceeding the initial pairwise estimate. The history term reduces the influence of edges with persistently large residuals, while the recovery term partially counteracts this reduction when an edge exhibits improved global consistency.

After the weight update, rotation and translation synchronization are recomputed using $w_{ij}^{(\ell+1)}$. This alternating process progressively transfers information between local pairwise reliability and global pose graph consistency. After $L$ IRLS iterations, the synchronized transformations $\{\mathbf{T}_i^{(L)}\}_{i=1}^{N}$ are taken as the final global poses.

The pairwise and synchronization stages complement each other. Spectral consistency focuses computation on reliable scan connections and geometrically coherent correspondence subsets, whereas adaptive history-aware reweighting further refines the resulting transformations according to global pose graph consistency. Together, they maintain a sparse and informative pose graph and improve registration robustness under low-overlap conditions.

\section{Experiments}
\label{sec:experiments}

\subsection{Experimental Setup}

\subsubsection{Datasets}
We evaluate GMPCR on four widely used datasets covering indoor, low-overlap, large-scale, and outdoor registration scenarios. \textbf{3DMatch}~\cite{zeng2017threedmatch} consists of indoor RGB-D reconstructions. We adopt its standard test split of eight scenes, containing 1,623 scan pairs with overlap ratios above $30\%$. \textbf{3DLoMatch}~\cite{huang2021predator} is constructed from the same test scenes but contains 1,781 more challenging scan pairs with overlap ratios between $10\%$ and $30\%$, providing a focused evaluation of registration robustness under limited overlap. \textbf{ScanNet}~\cite{dai2017scannet} contains large-scale indoor reconstructions. Following the common multiview registration protocol, we use 32 test scenes and sample 30 scans from each scene, yielding 960 scans and 13,920 evaluation pairs. \textbf{ETH}~\cite{pomerleau2012challenging} comprises four outdoor scenes with 132 scans and 713 official evaluation pairs. Its distinct sensing conditions, scene scales, and geometric structures enable us to assess the generalization capability of GMPCR from indoor to outdoor environments.

\subsubsection{Metrics}
For 3DMatch, 3DLoMatch, and ETH, Registration Recall (RR) is used as the primary metric, defined as the percentage of ground-truth scan pairs whose estimated relative transformations meet the given registration-error threshold. For ScanNet, rotation error (RE) and translation error (TE) are used to measure the difference between the estimated and ground-truth relative transformations. RE measures the angular difference between rotations, while TE measures the Euclidean distance between translation vectors. Threshold-based recall, together with the mean and median errors, is reported. Higher recall and lower errors indicate better registration performance.

\begin{table}[htbp]
    \centering
    \caption{Registration recall on 3DMatch, 3DLoMatch, and ETH.}
    \label{tab:registration_recall}
    \small
    \setlength{\tabcolsep}{6.5pt}
    \renewcommand{\arraystretch}{1.08}
    \begin{tabular}{@{}lccc@{}}
        \toprule
        & \multicolumn{3}{c}{Registration Recall (\%)} \\
        \cmidrule(l){2-4}
        & 3DMatch & 3DLoMatch & ETH \\
        \midrule
        EIGSE3~\cite{arrigoni2016spectral} & 40.1 & 26.5 & 96.3 \\
        L1-IRLS~\cite{chatterjee2013efficient} & 68.6 & 49.0 & 90.2 \\
        RotAvg~\cite{chatterjee2018robust} & 77.2 & 60.3 & 96.6 \\
        LITS~\cite{huang2019learning} & 80.8 & 65.2 & 48.4 \\
        HARA~\cite{lee2022hara} & 83.8 & 71.9 & 96.0 \\
        SGHR~\cite{wang2023robust} & 96.2 & 81.6 & 99.1 \\
        SMVR~\cite{fang2024robust} & 96.7 & 83.0 & 99.6 \\
        RAP~\cite{pan2026eccv} & 85.6 & 73.0 & \textbf{99.8} \\
        \midrule
        \textbf{GMPCR} & \textbf{97.2} & \textbf{89.6} & 99.7 \\
        \bottomrule
    \end{tabular}
\end{table}

\subsubsection{Implementation Details}
We compare GMPCR with EIGSE3~\cite{arrigoni2016spectral}, L1-IRLS~\cite{chatterjee2013efficient}, RotAvg~\cite{chatterjee2018robust}, LITS~\cite{huang2019learning}, HARA~\cite{lee2022hara}, LMVR~\cite{gojcic2020learning}, SGHR~\cite{wang2023robust}, SMVR~\cite{fang2024robust}, and RAP~\cite{pan2026eccv}. These methods cover classical and learning-based transformation synchronization, sparse pose graph construction, and correspondence-free multiview registration. For a fair comparison among pose graph-based methods, YOHO descriptors~\cite{wang2022yoho} are used for GMPCR and all synchronization baselines.

Unless otherwise specified, the dataset-dependent parameters follow the order of 3DMatch, 3DLoMatch, ScanNet, and ETH. We sample $N_s=\{200,240,200,400\}$ representative points from each scan and retain $N_o=\{100,100,100,75\}$ initial correspondences for each scan pair. The spectral retention ratios are set to $r_s=\{0.50,0.65,0.50,0.25\}$, and the neighborhood sizes are set to $K=\{6,8,8,4\}$. For each retained scan pair, at most $K_c=\{10,5,10,10\}$ confidence-ranked maximal cliques are used to generate transformation hypotheses. For adaptive history-aware synchronization, the maximum numbers of IRLS iterations are set to $L=\{10,90,80,5\}$. The memory-response parameters are set to $\kappa=\{3,1,5,20\}$, while the recovery coefficients are $\rho=\{0.1,0.1,0.1,0.5\}$. The experiments are conducted on a PC equipped with an Intel Core i5-14600KF CPU, 16GB RAM, and an NVIDIA GeForce RTX 5060Ti GPU.

\subsection{Qualitative and Quantitative Results}

As shown in Table~\ref{tab:registration_recall}, GMPCR achieves registration recalls of $97.2\%$ and $89.6\%$ on 3DMatch and 3DLoMatch, outperforming all competing methods. It exceeds SGHR by $1.0$ and $8.0$ percentage points, SMVR by $0.5$ and $6.6$ percentage points, and RAP by $11.6$ and $16.6$ percentage points on 3DMatch and 3DLoMatch, respectively. The advantage is particularly evident on 3DLoMatch, where low overlap leads to sparse reliable correspondences and increased matching ambiguity. These results demonstrate that the compatibility graph effectively strengthens geometrically consistent inliers while suppressing isolated or weakly connected outliers, enabling robust global registration under heavily contaminated correspondences. On ETH, RAP achieves the highest RR of $99.8\%$. GMPCR is only $0.1$ percentage points below the best result while outperforming SMVR and SGHR by 0.1 and 0.6 percentage points, respectively. Given the substantial differences in scene scale and sensing conditions between ETH and the indoor datasets, the consistently high performance indicates that GMPCR generalizes well across both indoor and outdoor scenarios. On ScanNet (Table~\ref{tab:scannet_results}), SMVR achieves the lowest mean rotation error of $18.9^{\circ}$, while GMPCR ranks second with $20.2^{\circ}$. GMPCR obtains a mean translation error of $0.46$m, matching SMVR and improving over SGHR at $0.56$m; it also achieves the lowest median translation error of $0.40$m. Compared with SGHR, GMPCR reduces the mean rotation error from $21.7^{\circ}$ to $20.2^{\circ}$ and the mean translation error from $0.56$m to $0.46$m. It also substantially outperforms RAP, whose mean rotation and translation errors reach $44.1^{\circ}$ and $1.05$m, respectively. These results show that GMPCR achieves competitive rotation accuracy together with strong translation accuracy on ScanNet.

\begin{table}[htbp]
    \centering
    \caption{Rotation and translation error on ScanNet.}
    \label{tab:scannet_results}
    \resizebox{\linewidth}{!}{ 
    \begin{tabular}{lccccc cc ccccc cc}
        \toprule
        & \multicolumn{7}{c}{Rotation error}
        & \multicolumn{7}{c}{Translation error (m)} \\
        \cmidrule(lr){2-8}\cmidrule(lr){9-15}
        
        & $3^\circ$ & $5^\circ$ & $10^\circ$ & $30^\circ$ & $45^\circ$
        & Mean$\downarrow$ & Med$\downarrow$
        & 0.05 & 0.10 & 0.25 & 0.50 & 0.75
        & Mean$\downarrow$ & Med$\downarrow$ \\
        \midrule
        EIGSE3~\cite{arrigoni2016spectral}
& 40.8 & 46.3 & 51.9 & 61.2 & 65.7
& $40.6^\circ$ & $37.1^\circ$
& 23.9 & 38.5 & 51.0 & 59.3 & 66.1 & 0.88 & 0.84 \\

L1-IRLS~\cite{chatterjee2013efficient}
& 46.3 & 54.2 & 61.6 & 64.3 & 66.8
& $41.8^\circ$ & $34.0^\circ$
& 24.1 & 38.5 & 48.3 & 55.6 & 60.9 & 1.05 & 1.01 \\

RotAvg~\cite{chatterjee2018robust}
& 50.2 & 60.1 & 65.3 & 66.8 & 68.8
& $38.5^\circ$ & $31.6^\circ$
& 31.8 & 49.0 & 58.8 & 63.3 & 65.6 & 0.96 & 0.83 \\

LITS~\cite{huang2019learning}
& 54.3 & 69.4 & 75.6 & 78.5 & 80.3
& $24.9^\circ$ & $19.9^\circ$
& 31.4 & 54.4 & 72.3 & 76.7 & 79.6 & 0.65 & 0.56 \\

HARA~\cite{lee2022hara}
& 55.7 & 63.7 & 69.0 & 70.8 & 72.1
& $34.7^\circ$ & $31.3^\circ$
& 35.2 & 53.6 & 65.4 & 68.6 & 71.7 & 0.86 & 0.71 \\

LMVR~\cite{gojcic2020learning}
& 48.3 & 53.6 & 58.9 & 63.2 & 64.0
& $48.1^\circ$ & $33.7^\circ$
& 34.5 & 49.1 & 58.5 & 61.6 & 63.9
& 0.83 & 0.55 \\

SGHR~\cite{wang2023robust}
& \textbf{59.1} & \textbf{73.1} & 80.8 & 82.5 & 83.0
& $21.7^\circ$ & $19.0^\circ$
& \textbf{39.9} & \textbf{64.1} & 76.7 & 79.0 & 81.9
& 0.56 & 0.49 \\

SMVR~\cite{fang2024robust}
& 58.0 & 71.4 & 80.0 & 86.0 & 86.5
& \bm{$18.9^\circ$} & \bm{$17.6^\circ$}
& 39.0 & 61.7 & 76.6 & 81.5 & \textbf{84.8}
& \textbf{0.46} & 0.44 \\

RAP~\cite{pan2026eccv}
& 33.1 & 41.0 & 49.5 & 60.1 & 64.4
& $44.1^\circ$ & $28.0^\circ$
& 20.9 & 33.2 & 46.8 & 56.1 & 62.0 & 1.05 & 0.85 \\

\midrule
GMPCR
& 54.8 & 70.9 & \textbf{83.4}
& \textbf{86.3} & \textbf{86.6}
& $20.2^\circ$ & $19.2^\circ$
& 35.4 & 59.7 & \textbf{79.4} & \textbf{82.8} & 84.4
& \textbf{0.46} & \textbf{0.40} \\
        \bottomrule
    \end{tabular}
    }
\end{table}

\begin{figure}[htbp]
  \centering
  \setlength{\tabcolsep}{2pt}
  \renewcommand{\arraystretch}{0}
  \begin{tabular}{@{}c@{\hspace{2pt}}cccc@{}}
    & \textbf{Ground Truth} & \textbf{SGHR} & \textbf{SMVR} & \textbf{GMPCR} \\[2pt]

    \parbox[c][0.22\textwidth][c]{1.4em}
      {\centering\rotatebox[origin=c]{90}{\textbf{Kitchen}}}
    & \raisebox{-0.5\height}{\includegraphics[width=0.22\textwidth]{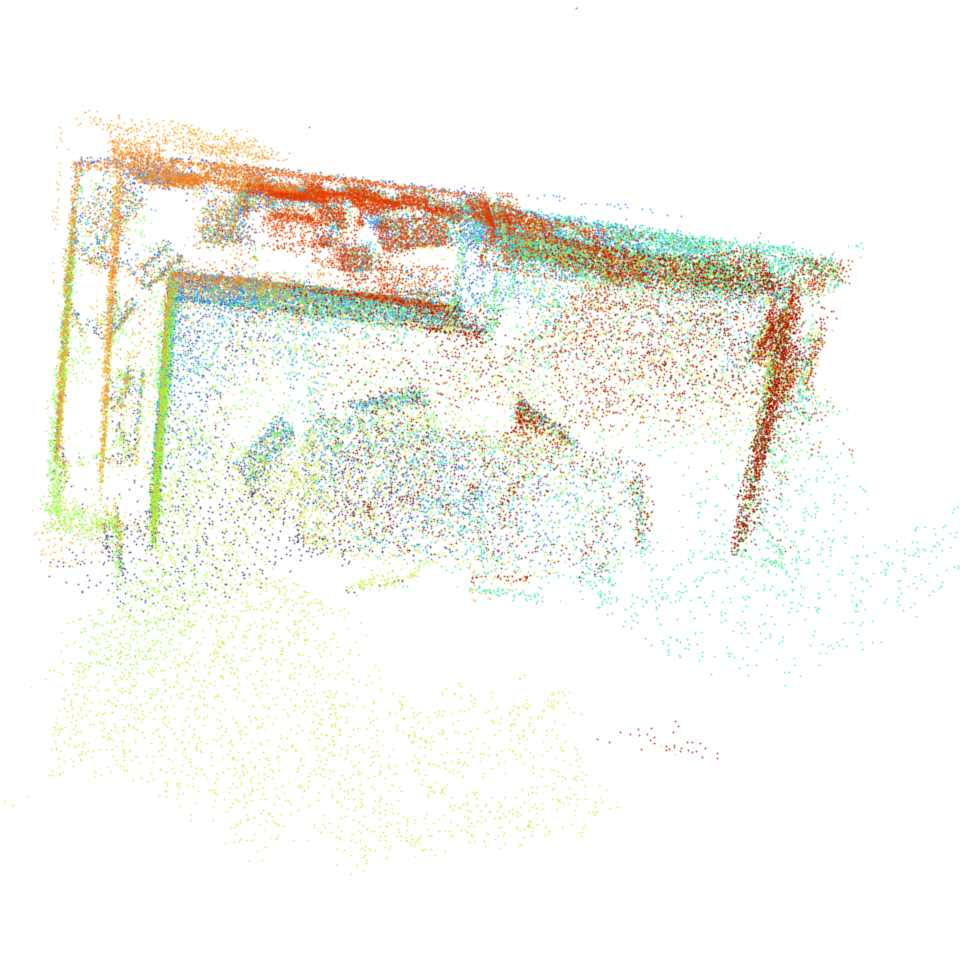}}
    & \raisebox{-0.5\height}{\includegraphics[width=0.22\textwidth]{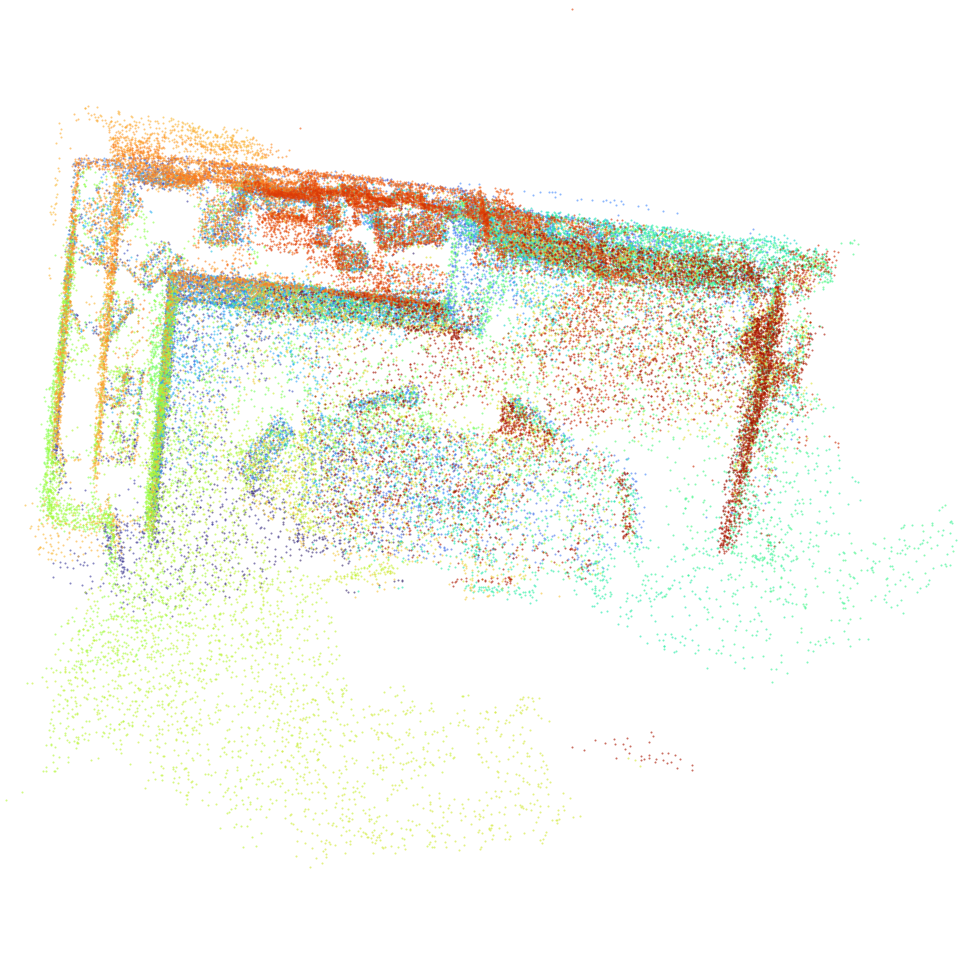}}
    & \raisebox{-0.5\height}{\includegraphics[width=0.22\textwidth]{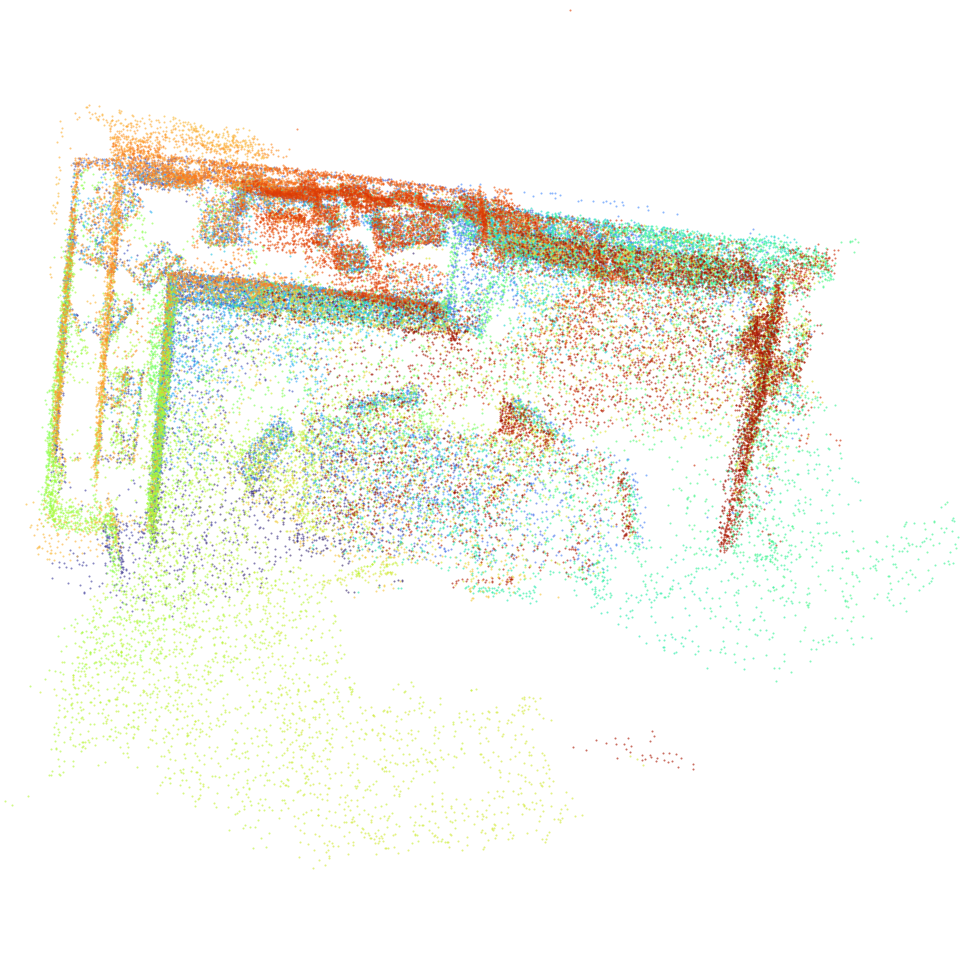}}
    & \raisebox{-0.5\height}{\includegraphics[width=0.22\textwidth]{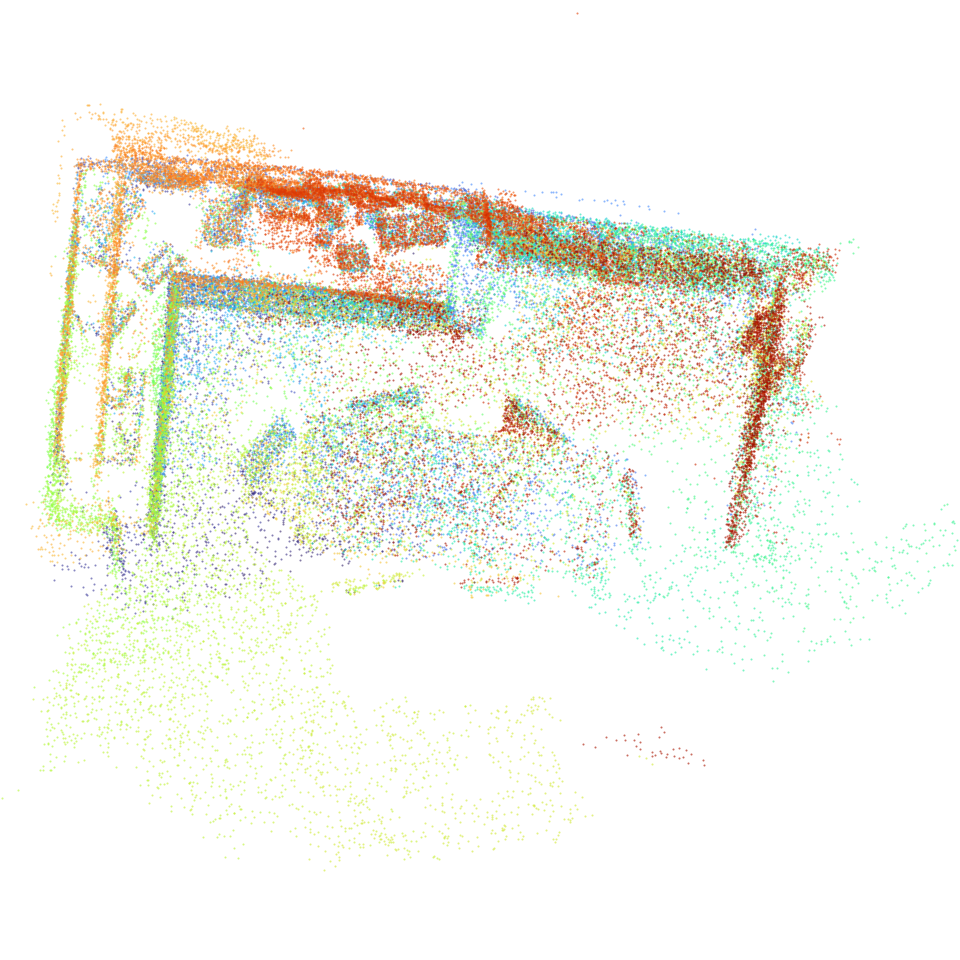}}
    \\[-1pt]

    \parbox[c][0.22\textwidth][c]{1.4em}
      {\centering\rotatebox[origin=c]{90}{\textbf{Home-MD}}}
    & \raisebox{-0.5\height}{\includegraphics[width=0.22\textwidth]{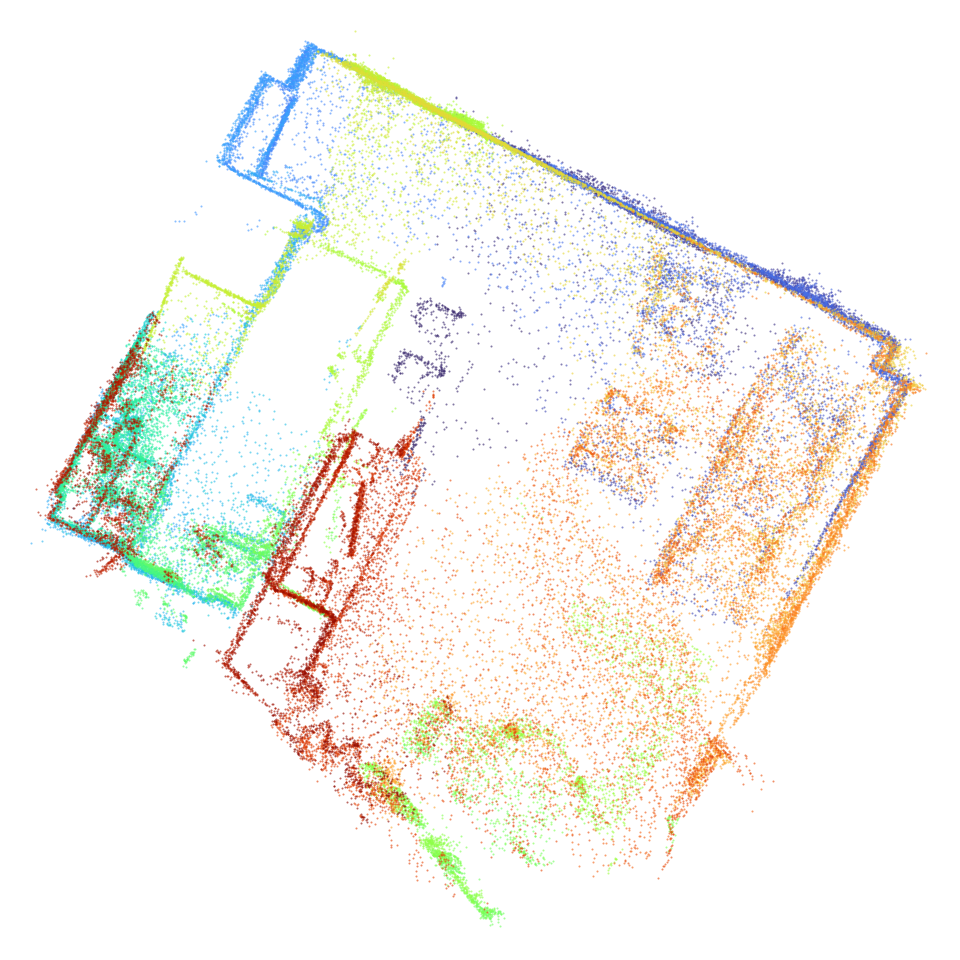}}
    & \raisebox{-0.5\height}{\includegraphics[width=0.22\textwidth]{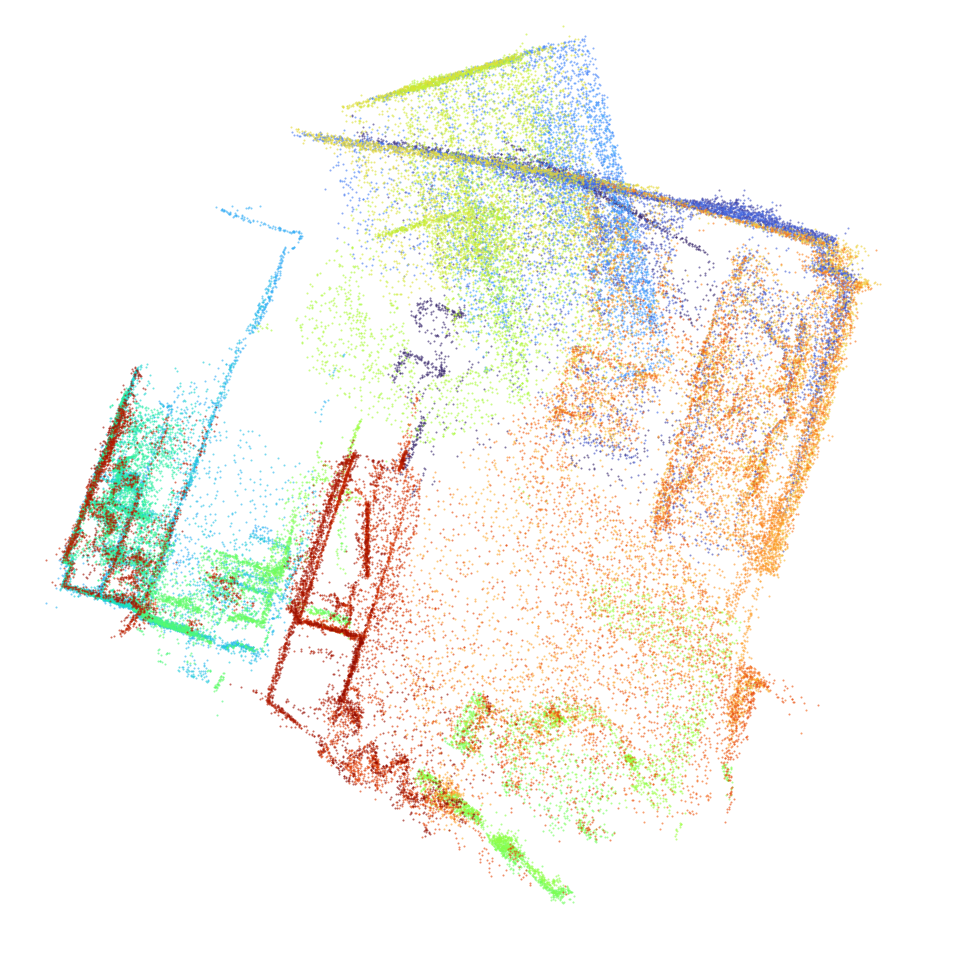}}
    & \raisebox{-0.5\height}{\includegraphics[width=0.22\textwidth]{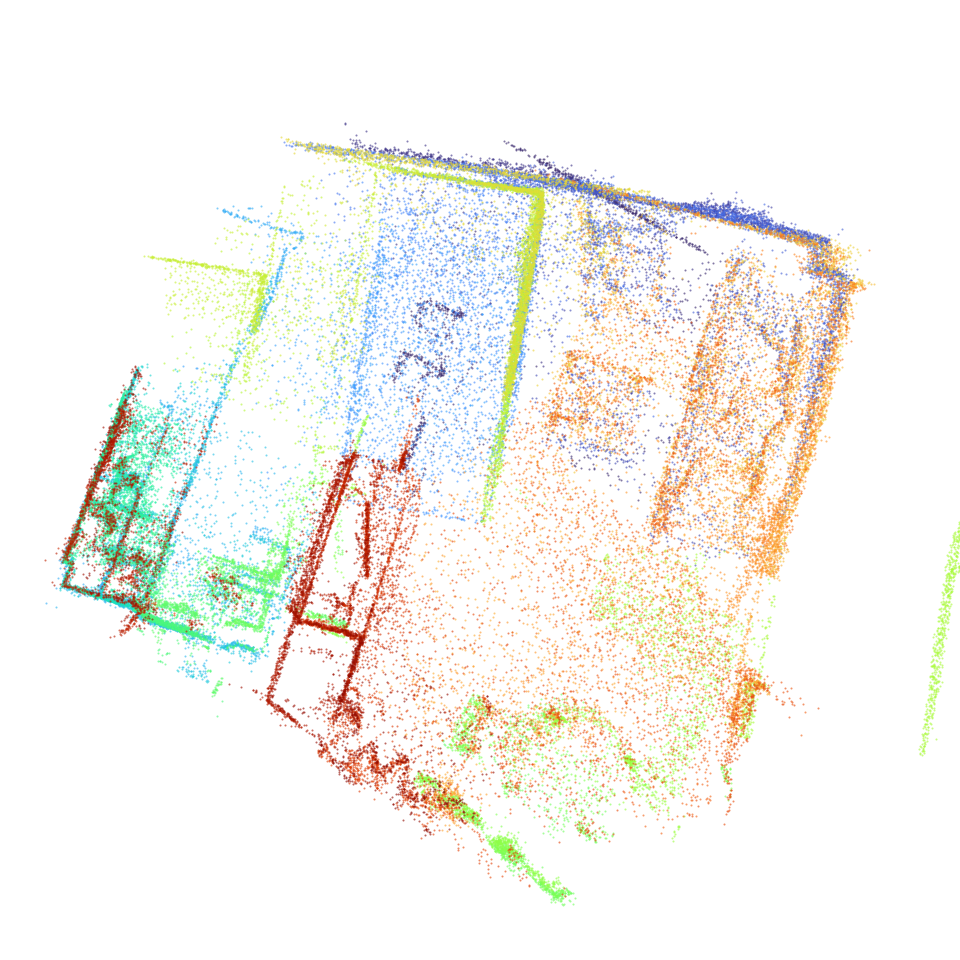}}
    & \raisebox{-0.5\height}{\includegraphics[width=0.22\textwidth]{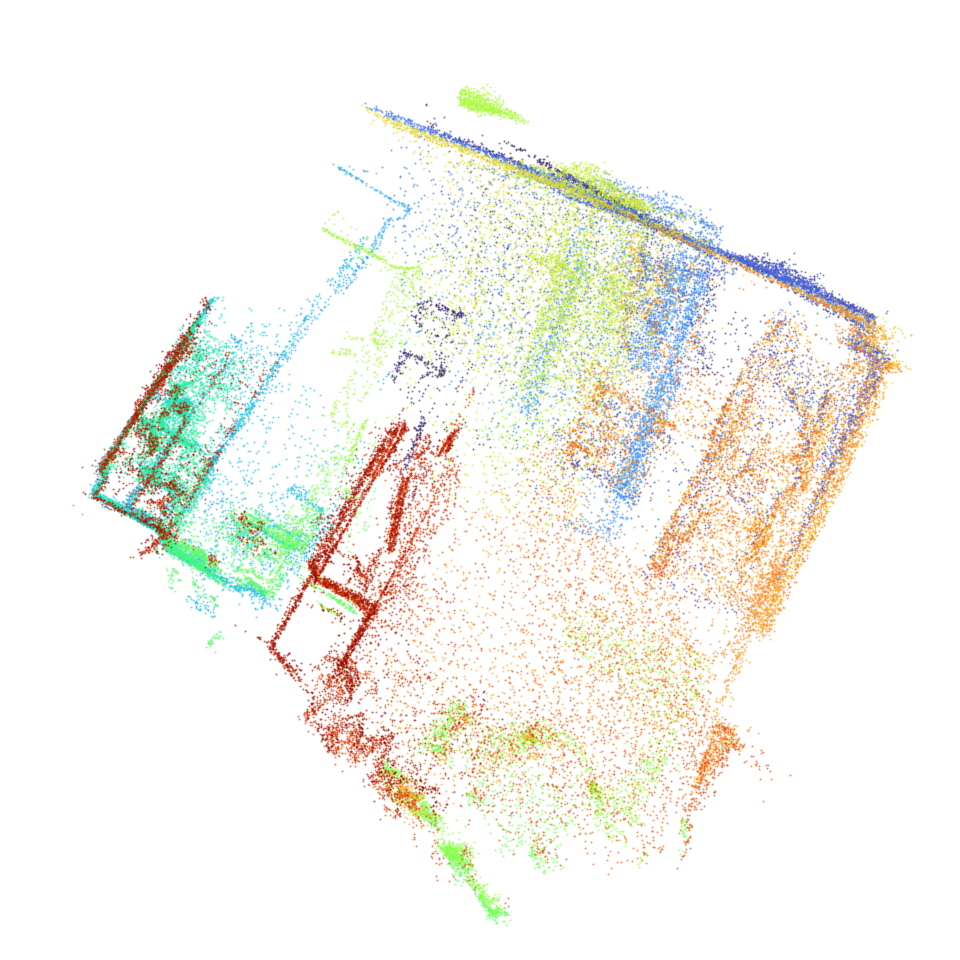}}
    \\[-1pt]

    \parbox[c][0.22\textwidth][c]{1.4em}
      {\centering\rotatebox[origin=c]{90}{\textbf{Studyroom}}}
    & \raisebox{-0.5\height}{\includegraphics[width=0.22\textwidth]{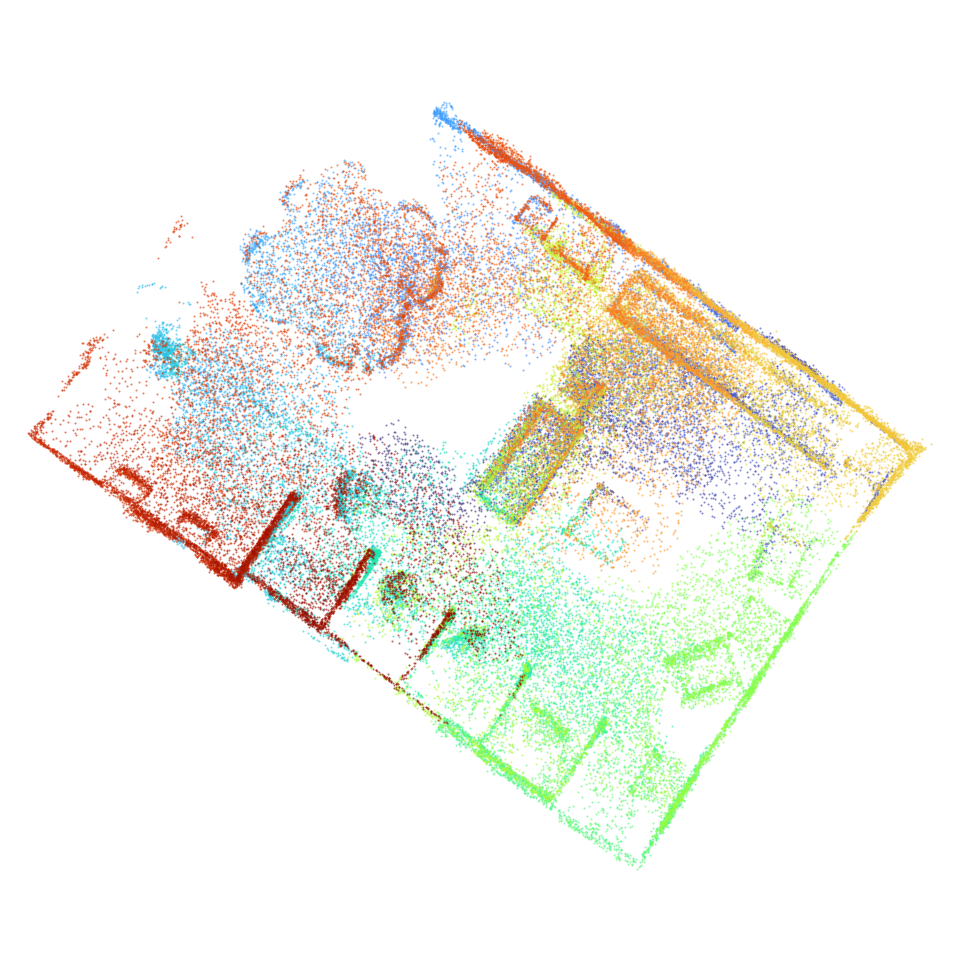}}
    & \raisebox{-0.5\height}{\includegraphics[width=0.22\textwidth]{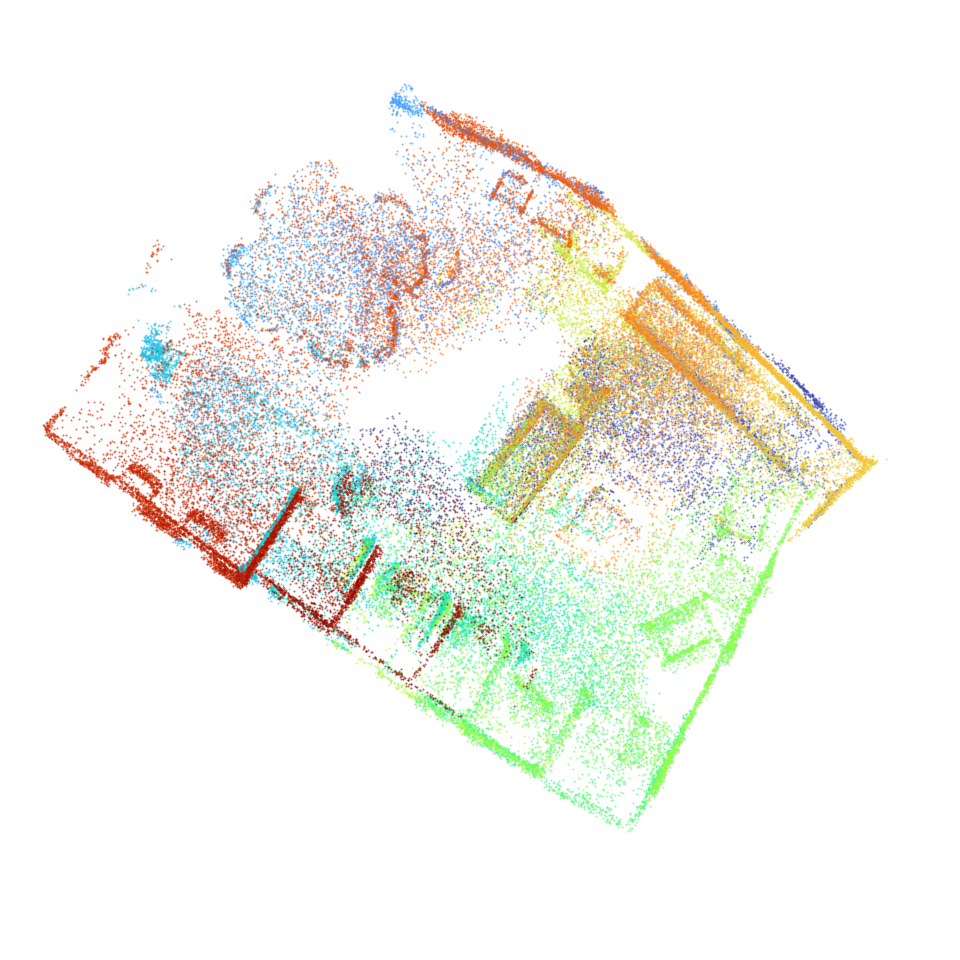}}
    & \raisebox{-0.5\height}{\includegraphics[width=0.22\textwidth]{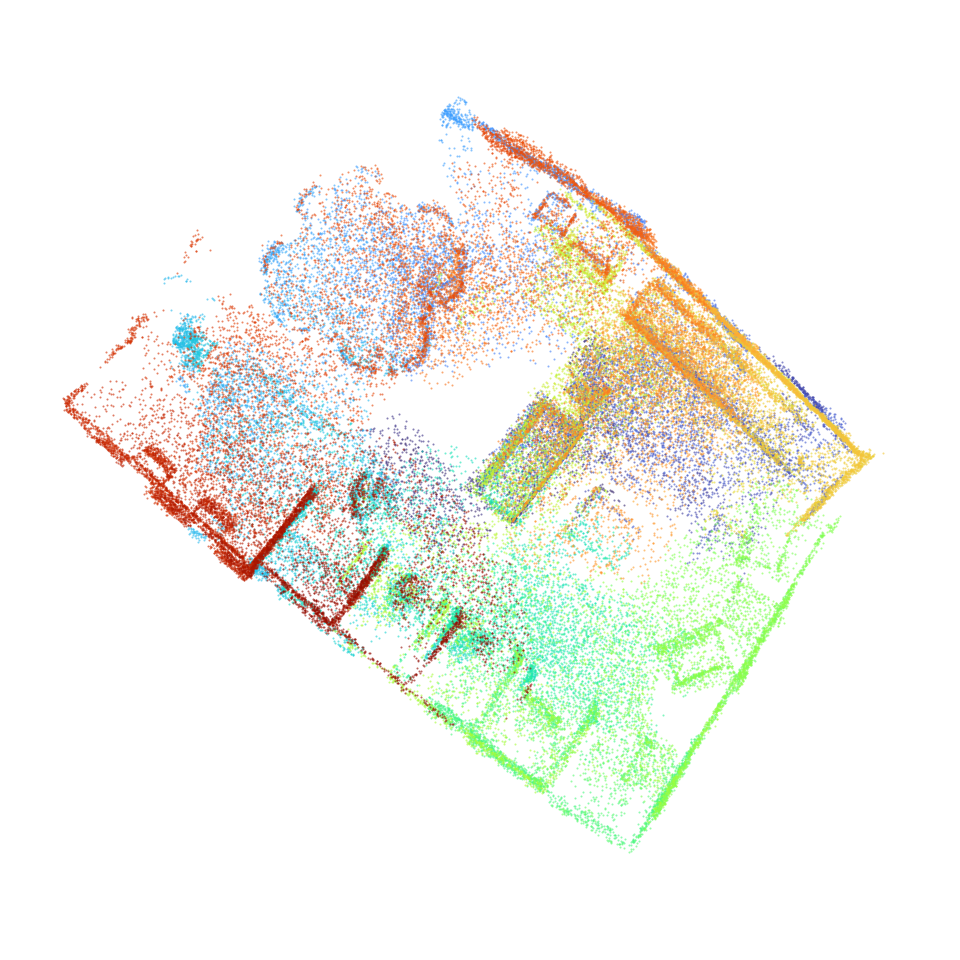}}
    & \raisebox{-0.5\height}{\includegraphics[width=0.22\textwidth]{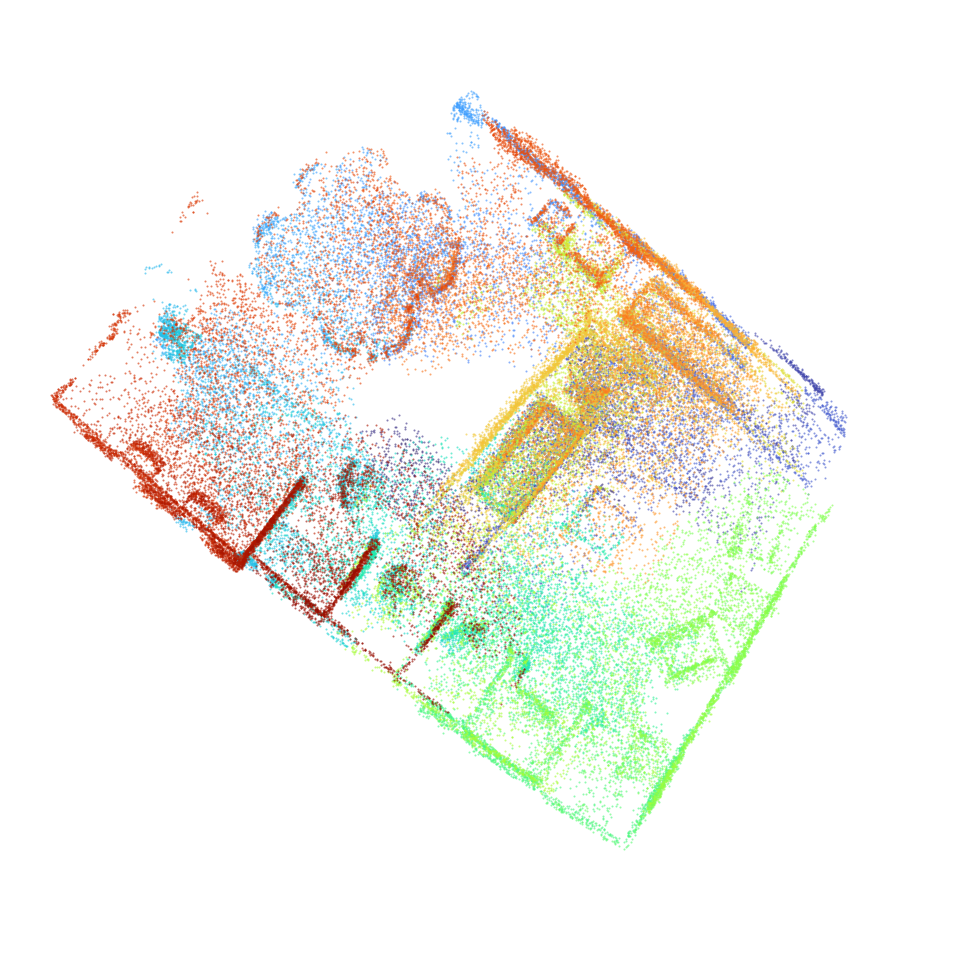}}
    \\[-1pt]

  \end{tabular}
  \caption{Qualitative comparison of global registration on 3DMatch. The leftmost column shows the ground truth, while the remaining columns show the results of SGHR, SMVR, and GMPCR.}
  \label{fig:global-registration-qualitative}
\end{figure}

Fig.~\ref{fig:global-registration-qualitative} compares the end-to-end global registration results of SGHR, SMVR, and GMPCR on three representative 3DMatch scenes. All methods recover coherent structures in the Kitchen scene. Their differences become more visible in Home-MD and Studyroom, which contain repetitive geometry and more complex spatial layouts. In these challenging scenes, SGHR and SMVR exhibit local boundary duplication and scan displacement, whereas GMPCR generally produces more consistent walls and room structures with less visible drift. These results qualitatively demonstrate the robustness of GMPCR against unreliable pairwise transformations in complex multiview registration scenarios.

Overall, GMPCR achieves the best performance on 3DMatch and 3DLoMatch while remaining competitive on ETH and ScanNet. Its clear advantage on 3DLoMatch demonstrates the effectiveness of geometric consistency under low-overlap conditions. Combined with spectral correspondence filtering and adaptive synchronization, GMPCR provides a favorable balance among registration accuracy, low-overlap robustness, and consistent performance across different datasets and scene types.

\subsection{Ablation Study}

\begin{table}[htbp]
\centering
\caption{Ablation study on 3DMatch and 3DLoMatch. All results are registration recall (\%).}
\label{tab:progressive_ablation}
\small
\setlength{\tabcolsep}{5pt}
\resizebox{\linewidth}{!}{%
\begin{tabular}{lcccccc}
\toprule
Configuration
& Pair selection
& Spectral filtering
& Memory coefficient
& Recovery
& 3DMatch $\uparrow$
& 3DLoMatch $\uparrow$ \\
\midrule
Baseline
& -- & -- & -- & --
& 91.6 & 82.5 \\

+ Pair Selection
& \checkmark & -- & -- & --
& 96.0 & 88.7 \\

+ Spectral Filtering
& \checkmark & \checkmark & -- & --
& 96.4 & 89.0 \\

+ Memory Coefficient
& \checkmark & \checkmark & \checkmark & --
& 97.0 & 89.5 \\

+ Recovery Term (Full)
& \checkmark & \checkmark & \checkmark & \checkmark
& \textbf{97.2} & \textbf{89.6} \\
\bottomrule
\end{tabular}}
\end{table}

\begin{figure}[htbp]
    \centering
    \begin{minipage}[t]{0.40\textwidth}
        \centering
        \includegraphics[width=\linewidth]{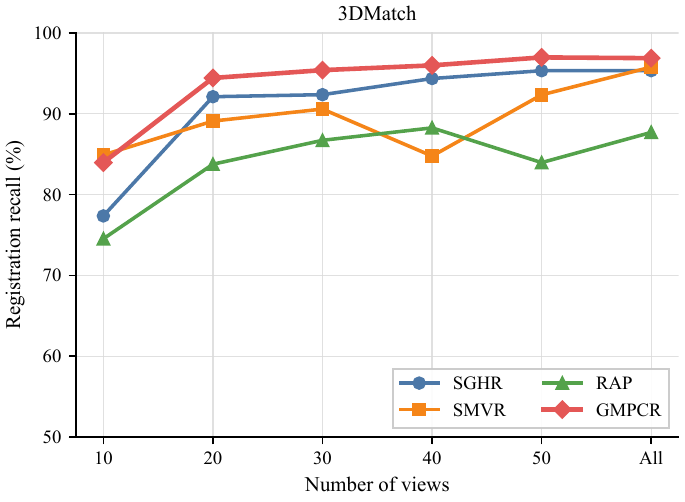}
        \par\textbf{(a)}
    \end{minipage}
    \hspace{0.03\textwidth}
    \begin{minipage}[t]{0.40\textwidth}
        \centering
        \includegraphics[width=\linewidth]{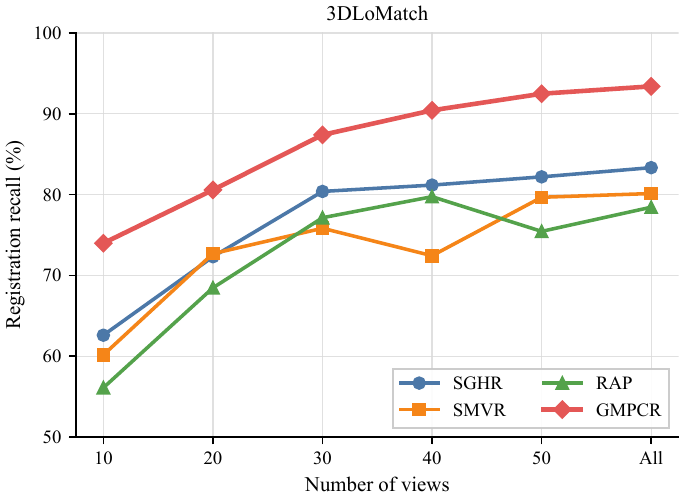}
        \par\textbf{(b)}
    \end{minipage}
    \caption{Registration recall under increasing numbers of jointly
    registered views on 3DMatch and 3DLoMatch.}
    \label{fig:view_scaling}
\end{figure}

\begin{table}[htbp]
    \centering
    \caption{Registration recall (\%) under increasing numbers of views on
    3DMatch and 3DLoMatch.}
    \label{tab:view_scaling_rr}
    \small
    \setlength{\tabcolsep}{4.2pt}
    \begin{tabular}{llrrrrrr}
        \toprule
        Dataset & Method & 10 & 20 & 30 & 40 & 50 & All \\
        \midrule
        3DMatch & SGHR~\cite{wang2023robust} & 77.4 & 92.1 & 92.4 & 94.4 & 95.3 & 95.3 \\
         & SMVR~\cite{fang2024robust} & \textbf{84.9} & 89.1 & 90.6 & 84.8 & 92.3 & 95.8 \\
         & RAP~\cite{pan2026eccv} & 74.5 & 83.8 & 86.7 & 88.3 & 84.0 & 87.7 \\
         & \textbf{GMPCR} & 84.0 & \textbf{94.4} & \textbf{95.4} & \textbf{96.0} & \textbf{97.0} & \textbf{96.9} \\
        \midrule
        3DLoMatch & SGHR~\cite{wang2023robust} & 62.6 & 72.3 & 80.4 & 81.2 & 82.2 & 83.3 \\
         & SMVR~\cite{fang2024robust} & 60.2 & 72.7 & 75.8 & 72.4 & 79.7 & 80.1 \\
         & RAP~\cite{pan2026eccv} & 56.1 & 68.5 & 77.1 & 79.7 & 75.4 & 78.4 \\
         & \textbf{GMPCR} & \textbf{74.0} & \textbf{80.6} & \textbf{87.4} & \textbf{90.4} & \textbf{92.5} & \textbf{93.4} \\
        \bottomrule
    \end{tabular}
\end{table}

We conduct an ablation study on 3DMatch and 3DLoMatch to evaluate the contribution of each GMPCR component. Starting from the baseline, we progressively introduce pair selection, spectral filtering, adaptive memory, and recovery while keeping other settings fixed. Without pair selection, all valid scan pairs are retained; without spectral filtering, all initial correspondences are used. Disabling adaptive memory replaces the convergence-dependent update with an SGHR-style history accumulation, while disabling recovery prevents down-weighted edges from regaining confidence. Thus, the baseline uses all scan pairs and correspondences, SGHR-style history update, and no recovery.

As shown in Table~\ref{tab:progressive_ablation}, the baseline achieves registration recalls of 91.6\% and 82.5\% on 3DMatch and 3DLoMatch. Pair selection increases them to 96.0\% and 88.7\%, giving the largest gains of 4.4 and 6.2 percentage points and demonstrating the importance of reliable pose graph construction. Spectral filtering further improves the recalls to 96.4\% and 89.0\% by removing inconsistent correspondences. Adaptive memory raises them to 97.0\% and 89.5\%, indicating that convergence-aware history weighting benefits global synchronization. Finally, recovery yields 97.2\% and 89.6\%, providing additional refinement by restoring reliable edges when their global consistency improves. Overall, all components contribute to performance, with pair selection providing the dominant improvement and the full model achieving the best results on both datasets.

\subsection{Robustness under increasing Numbers of Views}

We further evaluate the robustness of different methods under varying numbers of jointly registered views. Following RAP \cite{pan2026eccv}, we select six test scenes shared by 3DMatch and 3DLoMatch, each containing more than 50 views. For each selected scene, subsets with 10, 20, 30, 40, 50, and all available views are constructed by randomly sampling different numbers of views. Reducing the number of available views decreases the number of potential geometric connections and lowers the effective overlap among selected scans, making global registration more challenging. This experiment evaluates whether different methods can maintain reliable registration performance under sparse-view conditions and with increasing numbers of jointly registered views.

As shown in Table~\ref{tab:view_scaling_rr} and Fig.~\ref{fig:view_scaling}, GMPCR achieves the highest registration recall for all tested settings on 3DLoMatch and for 20 views and above on 3DMatch. On 3DMatch, GMPCR improves from 84.0\% with 10 views to 97.0\% with 50 views and maintains 96.9\% when all views are used. The advantage of GMPCR is more evident on 3DLoMatch, where increasing the number of views enlarges the registration problem and introduces more candidate scan relationships, including potentially unreliable low-overlap connections, while also providing richer geometric context. GMPCR achieves registration recalls of 74.0\%, 80.6\%, 87.4\%, 90.4\%, 92.5\%, and 93.4\% with increasing numbers of views, remaining consistently higher than all competing methods. With all available views, GMPCR improves over SGHR, SMVR, and RAP by 10.1, 13.3, and 15.0 percentage points, respectively. These results indicate that GMPCR can effectively handle larger multiview registration problems while maintaining stable registration performance. This stable improvement with increasing numbers of views is attributable to the reliable scan-pair selection and adaptive global synchronization, which allow the constructed pose graph to exploit additional geometric information while reducing the influence of unreliable connections. Overall, GMPCR demonstrates robust multiview registration performance under both standard and low-overlap conditions.

\begin{table}[htbp]
\centering
\caption{Scan-pair retrieval quality for pose graph construction at $K\in\{4,8,12\}$.}
\label{tab:pose_graph_quality}
\small
\setlength{\tabcolsep}{4.5pt}
\begin{tabular}{lclcc}
\toprule
Dataset & $K$ & Method & R@$K$ (\%) $\uparrow$ & Regret $\downarrow$ \\
\midrule
3DLoMatch & 4 & SGHR~\cite{wang2023robust} & 60.3 & 0.0866 \\
           & 4 & SMVR~\cite{fang2024robust} & 63.2 & 0.0820 \\
           & 4 & GMPCR & \textbf{73.2} & \textbf{0.0380} \\
\cmidrule(lr){2-5}
           & 8 & SGHR~\cite{wang2023robust} & 61.5 & 0.0749 \\
           & 8 & SMVR~\cite{fang2024robust} & 62.5 & 0.0699 \\
           & 8 & GMPCR & \textbf{74.5} & \textbf{0.0322} \\
\cmidrule(lr){2-5}
           & 12 & SGHR~\cite{wang2023robust} & 60.6 & 0.0617 \\
           & 12 & SMVR~\cite{fang2024robust} & 62.9 & 0.0552 \\
           & 12 & GMPCR & \textbf{72.7} & \textbf{0.0287} \\
\midrule
ScanNet & 4 & SGHR~\cite{wang2023robust} & 71.0 & 0.0664 \\
        & 4 & SMVR~\cite{fang2024robust} & 74.8 & 0.0485 \\
        & 4 & GMPCR & \textbf{78.0} & \textbf{0.0311} \\
\cmidrule(lr){2-5}
        & 8 & SGHR~\cite{wang2023robust} & 69.1 & 0.0694 \\
        & 8 & SMVR~\cite{fang2024robust} & 73.8 & 0.0498 \\
        & 8 & GMPCR & \textbf{77.7} & \textbf{0.0354} \\
\cmidrule(lr){2-5}
        & 12 & SGHR~\cite{wang2023robust} & 70.1 & 0.0576 \\
        & 12 & SMVR~\cite{fang2024robust} & 73.1 & 0.0461 \\
        & 12 & GMPCR & \textbf{76.7} & \textbf{0.0338} \\
\midrule
ETH & 4 & SGHR~\cite{wang2023robust} & 77.3 & 0.0235 \\
    & 4 & SMVR~\cite{fang2024robust} & 85.7 & 0.0133 \\
    & 4 & GMPCR & \textbf{86.4} & \textbf{0.0076} \\
\cmidrule(lr){2-5}
    & 8 & SGHR~\cite{wang2023robust} & 63.3 & 0.0276 \\
    & 8 & SMVR~\cite{fang2024robust} & 70.1 & 0.0209 \\
    & 8 & GMPCR & \textbf{79.0} & \textbf{0.0088} \\
\cmidrule(lr){2-5}
    & 12 & SGHR~\cite{wang2023robust} & 63.2 & 0.0247 \\
    & 12 & SMVR~\cite{fang2024robust} & 67.5 & 0.0188 \\
    & 12 & GMPCR & \textbf{77.9} & \textbf{0.0096} \\
\bottomrule
\end{tabular}
\end{table}

\subsection{Pose Graph Construction Quality}
\label{sec:pose_graph_quality}

We further evaluate the quality of the pose graphs constructed by different methods. A reliable pose graph should preserve scan pairs with high geometric overlap while avoiding unnecessary connections with weak constraints. Therefore, we compare the scan-pair rankings produced by GMPCR, SGHR, and SMVR using the ground-truth (GT) overlap between scans. Specifically, for each scan, the top-$K$ neighbors ranked by a method are compared with the GT top-$K$ neighbors that have the highest overlap ratios. We measure the ranking quality using Recall@K (R@K) and Regret. R@K is defined as the ratio of correctly retrieved high-overlap neighbors among the selected top-$K$ neighbors, where a higher value indicates that more informative scan connections are preserved. Regret measures the decrease in mean GT overlap relative to the GT top-$K$ neighbors, where a lower value indicates that the constructed pose graph is closer to the ideal graph.

As shown in Table~\ref{tab:pose_graph_quality}, GMPCR consistently achieves the best scan-pair retrieval performance across all datasets and different values of $K$. On 3DLoMatch, GMPCR obtains R@K values of 73.2\%, 74.5\%, and 72.7\% for $K=4$, $8$, and $12$, respectively, outperforming SGHR and SMVR by clear margins. Meanwhile, the corresponding Regret values are reduced to 0.0380, 0.0322, and 0.0287, indicating that GMPCR selects scan pairs closer to the highest-overlap neighbors. The improvements are particularly notable on 3DLoMatch, where low overlap makes reliable scan-pair selection more challenging. Similar trends are observed on ScanNet and ETH. On ScanNet, GMPCR achieves the highest R@K values of 78.0\%, 77.7\%, and 76.7\% under the three tested settings, together with the lowest Regret values. On ETH, GMPCR also consistently ranks the most informative scan connections, achieving R@K values of 86.4\%, 79.0\%, and 77.9\% for $K=4$, $8$, and $12$, respectively. These results show that the spectral consistency-based selection strategy can effectively identify reliable scan pairs across different scene types. Overall, the pose graph construction results demonstrate that GMPCR produces more accurate scan-pair rankings than existing sparse pose graph methods. By preserving informative geometric connections and reducing unreliable edges, GMPCR constructs a compact and reliable pose graph that provides stronger constraints for subsequent global pose synchronization.

\subsection{Pose Graph Connectivity Analysis}
\label{sec:pose_graph_connectivity}

We further analyze how the neighborhood size $K$ affects the connectivity of the constructed pose graph. For each scan, the top-$K$ scan pairs ranked by confidence are retained as neighbor candidates, and an undirected edge is added if either scan selects the other. We report the number of fully connected scenes (Conn.), the maximum number of connected components ($C_{\max}$), the minimum largest-connected-component ratio ($\mathrm{LCC}_{\min}$), the minimum reference-node reachability ratio ($R_{\min}$), and the mean number of retained edges ($\bar{M}$). These metrics jointly measure whether the selected scan pairs provide sufficient graph connectivity for subsequent pose synchronization.

\begin{table}[htbp]
    \centering
    \caption{Connectivity of pose graphs constructed by GMPCR under different 
    Top-$K$ settings. \emph{Conn.} is the number of connected scenes (out of
    eight), $C_{\max}$ is the maximum component count,
    $\mathrm{LCC}_{\min}$ and $R_{\min}$ are the minimum LCC and reachability
    ratios (\%), and $\bar{M}$ is the mean edge count.}
    \label{tab:GMPCR_pose_graph_connectivity_k}
    \small
    \renewcommand{\arraystretch}{1.08}
    \setlength{\tabcolsep}{5pt}
    \begin{tabular}{@{}c ccccc ccccc@{}}
        \toprule
        & \multicolumn{5}{c}{3DMatch}
        & \multicolumn{5}{c}{3DLoMatch} \\
        \cmidrule(lr){2-6}\cmidrule(lr){7-11}
        $K$
        & Conn. & $C_{\max}$ & $\mathrm{LCC}_{\min}$ & $R_{\min}$ & $\bar{M}$
        & Conn. & $C_{\max}$ & $\mathrm{LCC}_{\min}$ & $R_{\min}$ & $\bar{M}$ \\
        \midrule
        1  & $0/8$          & 19 & $8.8$   & $3.3$   & 39.0
           & $0/8$          & 18 & $8.3$   & $3.3$   & 39.0  \\
        2  & $3/8$          & 5  & $42.1$  & $42.1$  & 73.4
           & $1/8$          & 5  & $42.1$  & $6.7$   & 72.0  \\
        3  & $6/8$          & 3  & $57.9$  & $57.9$  & 106.9
           & $5/8$          & 3  & $45.9$  & $45.9$  & 105.1 \\
        4  & $6/8$          & 3  & $57.9$  & $57.9$  & 139.5
           & $6/8$          & 2  & $76.3$  & $76.3$  & 138.2 \\
        5  & $\mathbf{8/8}$ & 1  & $100.0$ & $100.0$ & 173.0
           & $7/8$          & 2  & $76.3$  & $76.3$  & 171.6 \\
        6  & $8/8$          & 1  & $100.0$ & $100.0$ & 206.2
           & $7/8$          & 2  & $76.3$  & $76.3$  & 203.8 \\
        7  & $8/8$          & 1  & $100.0$ & $100.0$ & 237.6
           & $7/8$          & 2  & $76.3$  & $76.3$  & 236.4 \\
        8  & $8/8$          & 1  & $100.0$ & $100.0$ & 271.6
           & $\mathbf{8/8}$ & 1  & $100.0$ & $100.0$ & 270.1 \\
        9  & $8/8$          & 1  & $100.0$ & $100.0$ & 304.6
           & $8/8$          & 1  & $100.0$ & $100.0$ & 303.6 \\
        10 & $8/8$          & 1  & $100.0$ & $100.0$ & 337.2
           & $8/8$          & 1  & $100.0$ & $100.0$ & 335.8 \\
        \bottomrule
    \end{tabular}
\end{table}

\begin{figure}[htbp]
    \centering
    \captionsetup{font=footnotesize}         
\captionsetup[subfloat]{font=scriptsize}  
    \subfloat[Connected scenes on 3DMatch.]{%
        \includegraphics[width=0.23\textwidth]{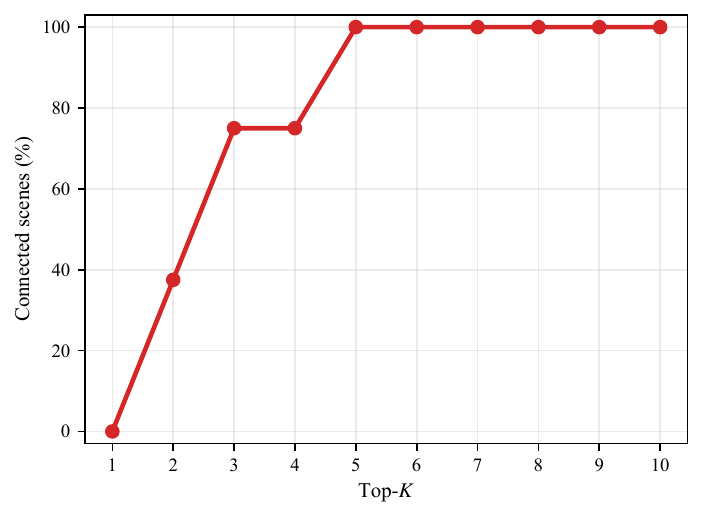}%
        \label{fig:GMPCR_connectivity_3dmatch_rate}%
    }
    \hfill
    \subfloat[Connected scenes on 3DLoMatch.]{%
        \includegraphics[width=0.23\textwidth]{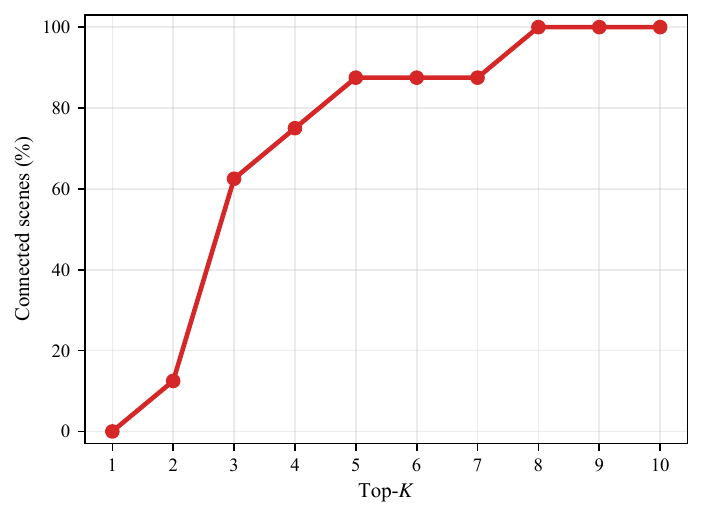}%
        \label{fig:GMPCR_connectivity_3dmatch_lcc}%
    }
    \hfill
    \subfloat[Worst-scene LCC on 3DMatch.]{%
        \includegraphics[width=0.23\textwidth]{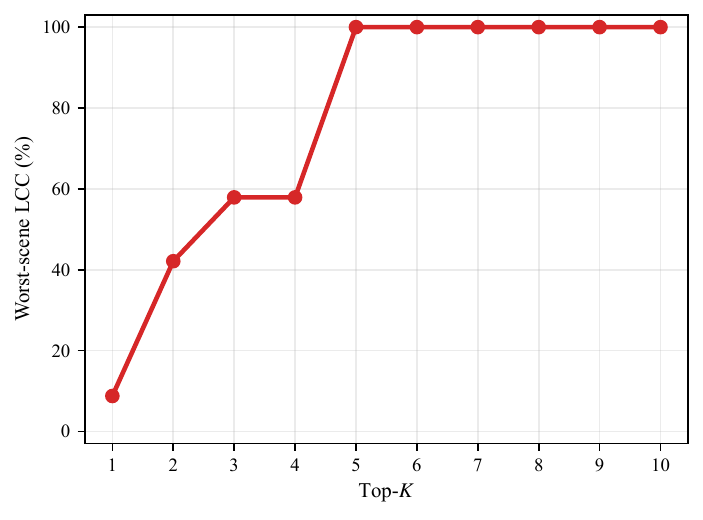}%
        \label{fig:GMPCR_connectivity_3dlomatch_rate}%
    }
    \hfill
    \subfloat[Worst-scene LCC on 3DLoMatch.]{%
        \includegraphics[width=0.23\textwidth]{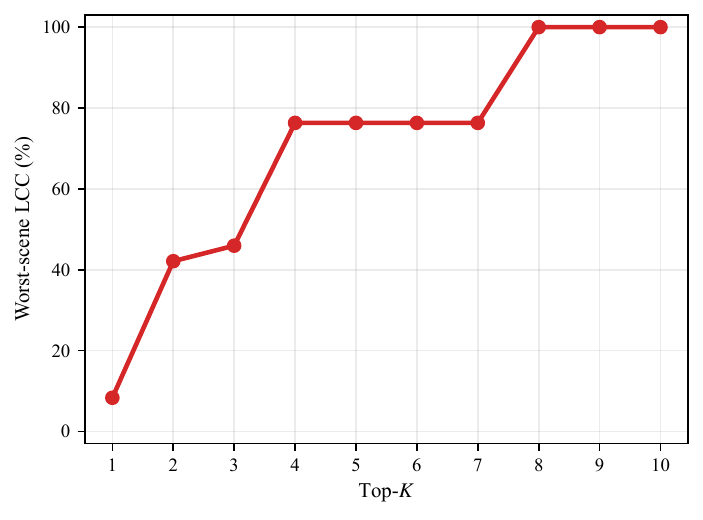}%
        \label{fig:GMPCR_connectivity_3dlomatch_lcc}%
    }
    \caption{Connectivity of pose graphs constructed by GMPCR as a function of $K$ on
    3DMatch and 3DLoMatch. Panels (a) and (b) show the percentage of fully
    connected scenes, while panels (c) and (d) report the LCC ratio of the
    worst scene.}
    \label{fig:pose_graph_connectivity_k}
\end{figure}

As shown in Table~\ref{tab:GMPCR_pose_graph_connectivity_k} and Fig.~\ref{fig:pose_graph_connectivity_k}, pose graph connectivity improves steadily as $K$ increases. On 3DMatch, no scene is fully connected at $K=1$, whereas six of eight scenes become connected at $K=3$. Full connectivity is achieved at $K=5$, where both $\mathrm{LCC}_{\min}$ and $R_{\min}$ reach 100\%, with an average of 173.0 edges. In contrast, 3DLoMatch requires a larger neighborhood because of its lower overlap. Seven scenes are connected for $K=5$--$7$, and full connectivity is reached at $K=8$, with $\mathrm{LCC}_{\min}=R_{\min}=100\%$ and 270.1 edges on average. These results also reveal a clear trade-off between connectivity and sparsity. Increasing $K$ improves connectivity but continuously increases the number of retained edges. Once full connectivity is reached, further increasing $K$ provides no connectivity benefit while making the graph denser. For example, increasing $K$ from 5 to 10 on 3DMatch raises the mean edge count from 173.0 to 337.2 without changing connectivity. Overall, confidence-based top-$K$ selection can construct sparse yet globally connected pose graphs, although lower-overlap scenes require more neighboring connections.

\subsection{Time Efficiency Analysis}

Table~\ref{tab:time_efficiency_3dmatch} reports the average runtime per scene on the eight 3DMatch test scenes. For SGHR, SMVR, and GMPCR, the runtime is divided into the main module and transformation synchronization. The main module includes graph construction and pairwise relative pose estimation, while the transformation synchronization time measures the cost of global pose optimization.

As shown in Table~\ref{tab:time_efficiency_3dmatch}, GMPCR achieves the lowest runtime among the pose graph-based methods, requiring only 23.7s per scene. It reduces the total runtime by 15.7s and 19.6s compared with SGHR and SMVR, corresponding to speedups of $1.66\times$ and $1.83\times$, respectively. The efficiency improvement is obtained from both stages of the pipeline. Specifically, GMPCR reduces the main-module runtime to 23.3s, which is 37.2\% and 38.8\% lower than SGHR and SMVR, respectively. Meanwhile, its transformation synchronization requires only 0.4s, achieving reductions of 82.6\% and 92.3\% compared with SGHR and SMVR. RAP achieves a lower total runtime of 10.4s, mainly benefiting from its feed-forward inference pipeline after training. In contrast, GMPCR performs 
\begin{table}[htbp]
    \centering
    \caption{Average runtime for registering one 3DMatch scene.}
    \label{tab:time_efficiency_3dmatch}
    \small
    \setlength{\tabcolsep}{3.8pt}
    \begin{tabular}{lrrr}
        \toprule
        Method & Main module (s) & Trans. Sync. (s) & Total (s) \\
        \midrule
        SGHR~\cite{wang2023robust} & 37.1 & 2.3 & 39.4 \\
        SMVR~\cite{fang2024robust} & 38.1 & 5.2 & 43.3 \\
        RAP~\cite{pan2026eccv} & -- & -- & \textbf{10.4} \\
        \textbf{GMPCR} & \textbf{23.3} & \textbf{0.4} & 23.7 \\
        \bottomrule
    \end{tabular}
\end{table}
scan-pair selection and global alignment using geometric consistency without requiring training. These two methods therefore follow different computational paradigms. Nevertheless, GMPCR achieves substantially lower runtime than existing pose graph-based methods while maintaining strong registration performance across different datasets and overlap conditions. The results demonstrate that GMPCR effectively reduces the computational cost of pose graph-based multiview registration, particularly in graph construction and transformation synchronization.

\section{Conclusion}

This paper presented GMPCR, a spectral consistency-guided framework for robust and efficient multiview point cloud registration in low-overlap scenes. GMPCR exploits a refined second-order compatibility structure to jointly evaluate correspondence reliability and scan-pair confidence, enabling informative scan pairs to be selected before relative transformation estimation and producing a sparse pose graph. For the retained pairs, clique-based hypothesis generation estimates reliable relative transformations. The pose graph is then refined by adaptive history-aware synchronization, which adjusts residual-history contributions according to global rotation consistency and allows previously down-weighted edges to recover when their consistency improves. GMPCR therefore integrates correspondence-level, scan-pair-level, and global pose consistency within a unified local-to-global framework.

Experiments on 3DMatch, 3DLoMatch, ScanNet, and ETH demonstrate its effectiveness. GMPCR achieves registration recalls of 97.2\% and 89.6\% on 3DMatch and 3DLoMatch, respectively, while remaining competitive on ScanNet and ETH. Scan-pair retrieval and runtime analyses further confirm that GMPCR identifies informative connections more accurately and operates more efficiently than existing sparse pose graph methods. Future work will focus on reducing candidate-pair evaluation costs and improving robustness in extremely sparse or weakly connected pose graphs.

\bibliographystyle{unsrt}
\bibliography{ref}

\end{document}